\documentclass{article}
\usepackage{iclr2027_conference,times}

\usepackage{amsmath,amsfonts,bm}

\def\eqref#1{equation~\ref{#1}}

\def\1{\bm{1}}

\DeclareMathAlphabet{\mathsfit}{\encodingdefault}{\sfdefault}{m}{sl}
\SetMathAlphabet{\mathsfit}{bold}{\encodingdefault}{\sfdefault}{bx}{n}

\usepackage{hyperref}
\usepackage{url}
\usepackage[utf8]{inputenc} 
\usepackage[T1]{fontenc}    
\usepackage{booktabs}       
\usepackage{amsfonts}       
\usepackage{nicefrac}       
\usepackage{verbatim}       
\usepackage{microtype}      
\usepackage{xcolor}         
\usepackage[normalem]{ulem}  
\usepackage{graphicx} 
\usepackage{tabularx}
\usepackage{amsmath}
\usepackage{amssymb}
\usepackage{multirow} 
\usepackage{svg}
\usepackage{wrapfig}
\usepackage{tcolorbox}
\usepackage{listings}
\tcbuselibrary{listings, breakable, skins}
\usepackage{float}       
\usepackage{placeins}
\usepackage{todonotes}
\usepackage{marginnote}
\usepackage{etoolbox}
\newcommand{\rev}[1]{\textcolor{black}{#1}}
\usepackage{enumitem}

\newtoggle{lmargin}
\newcommand{\altodo}[2][]{%
  \iftoggle{lmargin}{%
    \todo[#1]{#2}%
    \togglefalse{lmargin}%
  }{%
    {\let\marginpar\marginnote
     \reversemarginpar
     \todo[#1]{#2}}%
    \toggletrue{lmargin}%
  }%
}

\newcommand{\mdgym}[0]{\textsc{MDGym}{}}
\newcommand{\lammps}[0]{\textsc{Lammps}{}}
\newcommand{\gromacs}[0]{\textsc{Gromacs}{}}

\title{\mdgym: Benchmarking AI Agents on Molecular Simulations}

\author{
  Vinay Kumar$^1$, Satyendra Rajput$^2$, Mausam$^{1,2}$, N. M. Anoop Krishnan$^{1,3}$ \\
  $^1$ Yardi School of Artificial Intelligence, Indian Institute of Technology Delhi \\
  $^2$ Department of Computer Science and Engineering, Indian Institute of Technology Delhi \\
  $^3$ Department of Civil and Environmental Engineering, Indian Institute of Technology Delhi
}

\iclrfinalcopy 
\begin{document}

\maketitle
{\renewcommand{\thefootnote}{}%
 \footnotetext{\href{mailto:vinay.kumar@scai.iitd.ac.in}{\texttt{vinay.kumar@scai.iitd.ac.in}}, \href{mailto:msz208502@iitd.ac.in}{\texttt{msz208502@iitd.ac.in}}, \href{mailto:mausam@iitd.ac.in}{\texttt{mausam@iitd.ac.in}},
   \href{mailto:krishnan@iitd.ac.in}{\texttt{krishnan@iitd.ac.in}}.}}

\begin{abstract}
The promise of AI-driven scientific discovery hinges on whether AI agents can autonomously design and execute the computational workflows that underpin modern science. Molecular dynamics (MD) simulation provides a demanding test of this capability: it requires translating a physical problem into syntactically and semantically correct input scripts, choosing force fields, ensembles, and initial and boundary conditions, diagnosing engine errors and unstable trajectories, and interpreting outputs against known physical behavior. We introduce \mdgym{}, a benchmark of 200 expert-curated MD simulation problems in \lammps{} and \gromacs{}, two widely used MD engines, organized into three difficulty levels and comprising 366 target quantities scored against reference values from expert simulations. We evaluate three general-purpose command-line agent frameworks, Claude Code, Codex, and OpenHands, each with four LLMs. The best configuration solves \rev{69}\% of easy problems, 30\% of medium problems, and \rev{30\%} of hard problems, and the choice of model changes performance far more than the choice of framework. Trajectory and engine-error analyses show that most engine errors arise while agents assemble the simulation input, that weaker models fail early through syntax errors and premature termination, and that stronger models complete more simulations but are physically incorrect.
Current agents therefore lack three capabilities required for autonomous simulation: coherent specification of simulation inputs, domain-aware recovery from engine errors, and physical and methodological validation of results.

\end{abstract}

\section{Introduction}

Autonomous scientific discovery, in which AI systems formulate hypotheses, design experiments, execute computational workflows, and interpret results against physical theory, has become a central goal of AI research~\citep{wang_2023scientific,lu2024ai,Boiko_2023}. Agents operating in scientific domains must choose an appropriate methodology, design experiments grounded in the relevant physical concepts, and interpret numerical outputs against known physical laws to arrive at meaningful conclusions. Correctness in such tasks is defined by physical validity, which compilers and test suites for software do not verify~\citep{bisht2026agentic}. Determining whether current AI agents possess these capabilities, and where they fall short, is therefore a prerequisite for a credible path toward autonomous scientific discovery. 


Molecular dynamics (MD) simulation,
a method for simulating the behavior of materials and molecules at the atomic scale, is central to drug discovery~\citep{ahmed2023byte}, materials design~\citep{choi2023review}, and biophysics~\citep{sinha2022applications}. A successful MD simulation requires converting a scientific problem stated in natural language into a physically grounded model, justifying each modeling choice against the real system, and analyzing the results with methods appropriate to the relevant length and time scales. For systems of realistic complexity, the outcome of the simulation, such as the equilibrium density of a polymer melt or the diffusion coefficient of ions in solution, is unknown until the simulation is run. MD simulations therefore resemble laboratory experiments more closely than standard coding tasks and are often referred to as \textit{in silico} experiments.

An agent performing MD must translate a physical description of a system into syntactically and semantically correct input scripts, reason about initial and boundary conditions, select appropriate force fields and integration parameters, monitor trajectories for numerical instability, and interpret output quantities against known physical behavior. These competencies must be composed across a multi-stage pipeline of system setup, initialization, energy minimization, equilibration, production, and post-processing. A physically incoherent choice at any stage can yield a simulation that runs to completion and produces output without raising any error, yet whose result is physically meaningless. Since the correct outcome is unknown \emph{a priori}, the agent must detect and recover from such errors through scientific reasoning alone. The combination of long-horizon, multi-stage execution with domain knowledge makes MD simulation a demanding setting to examine a question \rev{that comes before autonomous discovery: can AI agents correctly setup and execute a scientific experiment once its physical model is specified?}



Existing benchmarks for MD agents address parts of this question. They evaluate small task sets, target a single engine or a single class of systems, or evaluate whether codes are executed without comparing the simulated quantities against reference values (Section~\ref{sec:related}). In response, we introduce \textbf{\mdgym}, an expert-curated benchmark of 200 MD simulation problems in two widely used engines, \lammps{}~\citep{Thompson_LAMMPS} and \gromacs{}~\citep{Hess_2008gromacs}, organized into three levels of increasing difficulty. The problems cover diverse types of materials such as biomolecules, metals and alloys, polymers, and carbon-based nanomaterials in crystalline and amorphous solid, liquid, and gaseous states. 
Each problem reflects a scientific question encountered in a research setting, is developed and verified by domain experts, and is scored automatically against physically meaningful reference quantities. \rev{\mdgym{} therefore evaluates the judgment needed to turn a stated scientific question into a correct measurement, but not the choice of the physical model. We treat this as the first requirement for autonomous simulation, since an agent that cannot reproduce a specified simulation cannot be trusted to design one.} Using \mdgym, we evaluate three general-purpose command-line agent frameworks, Claude Code, Codex, and OpenHands, each instantiated with four LLMs spanning frontier proprietary and open-weight models. Our contributions are as follows.
\begin{itemize}
    \item \textbf{Benchmark:} We release \mdgym{}, 
    comprising 200 simulation problems with 366 target quantities across 76 physical properties and 160 knowledge questions, together with a modular evaluation harness that extends to additional agents and MD engines, for tracking progress toward autonomous scientific computing.
    \item \textbf{Evaluation:} We evaluate 12 agent--model configurations on \mdgym{}. The best configuration solves \rev{69}\% of easy problems and 30\% of medium and \rev{30}\% of hard problems; the choice of model changes performance far more than the choice of agent framework.
    \item \textbf{Diagnosis:} We present a taxonomy of failure modes, derived from trajectory analysis and engine-level error classification, that localizes where in the simulation pipeline current agents fail. Our taxonomy identifies the capabilities that agents must acquire before they can serve as reliable partners in computational science.
\end{itemize}

\section{Related work}
\label{sec:related}

MD simulation integrates Hamilton's equations of motion for an atomistic system, with interatomic forces derived from an empirical force field that approximates the quantum mechanical potential energy surface. Thermostats and barostats extend the microcanonical (NVE) ensemble to constant-temperature and constant-pressure conditions, and non-equilibrium protocols such as uniaxial loading, planar shear, or imposed thermal gradients yield mechanical and transport properties. Physical validity requires a timestep that resolves the fastest vibrational mode, a force field applied within its domain of parameterization, ensemble and coupling constants that match the target conditions, and an analysis procedure appropriate to the measured quantity. Refer to App.~\ref{app:md_protocol} for a discussion on MD simulations. 

\textbf{Benchmarks for coding agents.} Since MD simulation scripts are written in specialized programming languages, our task is related to other coding tasks. Popular benchmarks for code generation and repository-level software engineering, such as HumanEval-V~\citep{zhang2024humaneval} and SWE-bench~\citep{jimenez_2023swe}, have largely been saturated. Terminal-Bench~\citep{merrill_2026terminal} evaluates agents on 89 realistic terminal tasks spanning scientific computing, software engineering, and system administration.
SUPER~\citep{bogin_2024} evaluates the setup and execution of 45 end-to-end problems from ML and NLP research repositories, and SciReplicate-Bench~\citep{xiang2025scireplicate} the reproduction of algorithms from research papers. AInsteinBench~\citep{duston_2025ainsteinbench} extends this line to scientific repositories and finds that agents struggle to preserve scientific invariants and maintain correctness in complex scientific algorithms, and DevOps-Gym~\citep{tang_2026devops} covers the software DevOps cycle. In these benchmarks, correctness is syntactic or behavioral, and is verified by test suites, stack traces, or output matching, so errors surface as signals on which an agent can act; correctness in MD simulation is physical and lacks such a verifier.

\textbf{Benchmarks for scientific AI agents.} A growing body of benchmarks evaluates LLM agents on scientific tasks (Table~\ref{tab:tasks_comparison}); App.~\ref{app:ai_agents} reviews AI scientist systems. ScienceAgentBench~\citep{chen_2024scienceagentbench} draws its tasks from peer-reviewed publications in four disciplines. SciCode~\citep{tian2024scicode} decomposes research-level problems from 16 natural-science subfields into subproblems and evaluates LLMs on these problems. BixBench~\citep{mitchener2025bixbench} analyzes frontier models on bioinformatics analysis scenarios in the open-answer regime.  AILA~\citep{mandal_evaluating_2025} evaluates LLM agents that operate an atomic force microscope in live laboratory experiments, and LAB-Bench~\citep{laurent_2024lab} tests practical biology research through over 2,400 multiple-choice questions on literature reasoning, database navigation, and sequence manipulation. More recently, \citet{rios2026ai} evaluated AI agents across eight environments and more than 25,000 runs. While earlier works focused on QA abilities of LLMs, recent works evaluate their agentic  abilities to solve scientific questions or a part thereof. \mdgym{} contributes to this family of work by framing MD simulation as an end-to-end problem to be solved by agents.

\textbf{Agents and benchmarks for MD simulation.} Existing agents for MD target specific systems or workflows. MDCrow~\citep{campbell_2026mdcrow} automates protein simulation tasks through a predefined set of tools; DynaMate~\citep{guilbert_2025dynamate} is a multi-agent framework for end-to-end workflows on protein and protein--ligand systems; PolyJarvis~\citep{zhao2026polyjarvis} couples an LLM with the RadonPy platform to predict polymer properties; and NAMD-Agent~\citep{chandrasekhar_2025automating} automates protein simulations in NAMD. These agents are evaluated on small task sets defined by their authors, ranging from 4 to 40 problems for the MD benchmarks in Table~\ref{tab:tasks_comparison}.The closest prior work, MDAgent2~\citep{shi2026mdagent2}, extends MDAgent~\citep{shi2025fine} with a multi-agent system for \lammps{} code generation, execution, and self-correction, and introduces MD-EvalBench for \lammps{} code generation and question answering. MD-EvalBench covers a single engine and scores each generated script by whether it executes and by expert ratings of the script, without comparing the simulated quantities against reference values. \mdgym{} addresses these gaps with 200 expert-curated problems across \lammps{} and \gromacs{} at three difficulty levels, each scored against physically meaningful reference quantities (Table~\ref{tab:tasks_comparison}), together with a modular harness that extends to additional engines such as AMBER, NAMD, and OpenMM. Since every problem requires executing simulations to completion, this scale entails a substantial computational cost (Sec.~\ref{sec:mdgym}).

\begin{table}[t]
\centering
\small
\vspace{-0.2in}
\caption{Benchmarks (excluding QA) for scientific AI agents that require code generation, simulation, or experimentation. 
Counts follow each benchmark's own definition of a problem.}
\label{tab:tasks_comparison}
\begin{tabularx}{\linewidth}{@{}l X r@{}}
\toprule
\textbf{Benchmark} & \textbf{Task type} & \textbf{Problems} \\
\midrule
QUASAR~\citep{yang2026quasar} & Atomistic simulation (DFT, MLP, MD, MC) & 9 \\
DynaMate~\citep{guilbert_2025dynamate} & Protein and protein--ligand MD workflows & 12 \\
MDCrow~\citep{campbell_2026mdcrow} & Protein MD through predefined tools & 25 \\
MD-EvalBench~\citep{shi2026mdagent2} & \lammps{} code generation & 566 \\
\midrule
SUPER~\citep{bogin_2024} & Executing ML and NLP research repositories & 45 \\
BixBench~\citep{mitchener2025bixbench} & Bioinformatics data analysis & 53 \\
SciCode~\citep{tian2024scicode} & Research coding in the natural sciences & 80 \\
AILA~\citep{mandal_evaluating_2025} & Atomic force microscopy experiments & 100 \\
ScienceAgentBench~\citep{chen_2024scienceagentbench} & Data-driven scientific discovery & 102 \\
\midrule
\textbf{\mdgym{} (ours)} & \textbf{End-to-end MD in \lammps{} and \gromacs{}} & \textbf{200} \\
\bottomrule
\vspace{-0.2in}
\end{tabularx}
\end{table}

\vspace{-0.05in}
\section{\mdgym{} Dataset \& Evaluation Harness}
\vspace{-0.05in}
\label{sec:mdgym}


\mdgym{} comprises 200 simulation problems and 160 knowledge QA tasks, developed and curated by researchers with expertise in MD simulation. The problems are drawn from research publications and tutorials or constructed from scratch. \gromacs{} problems adapt input files from the \gromacs{} repository,\footnote{\url{https://www.gromacs.org/}} published studies~\citep{rajput2023ethylene,rajput_2025}, and tutorials~\citep{Lemkul_2018,gravelle2025tutorials}; \lammps{} problems adapt examples from the official \lammps{} repository~\citep{Thompson_LAMMPS} and tutorials~\citep{gravelle2025tutorials}. Every sourced file is modified, for instance by substituting a different material, ensemble, or loading condition, so that no problem reduces to running an existing example verbatim, and each problem poses a research-level question. The knowledge questions are built on the problem statements and test syntax, such as correct command usage, and concepts, such as physical reasoning about ensembles and force fields (App.~\ref{app:dataset_details}).


Most problems follow a common protocol of energy minimization, NVT and/or NPT equilibration, and a production run from which the observables are extracted by post-processing; the harder problems additionally apply non-equilibrium protocols such as deformation, shear, or tensile loading. Reference values are obtained by executing expert-written simulation scripts in \gromacs{} and \lammps{} and computing the observables with Python scripts or with the analysis tools native to each engine. Each problem is stored as a JSON specification (App.~\ref{app:json}) that holds the problem description, difficulty level, engine, required metrics, reference values, and time limit; the orchestrator generates the agent prompt from this specification and passes the reference values to the validator.


\textbf{Quality control.} Problem curation and difficulty labelling were performed by a postdoctoral researcher with expertise in MD simulation and verified by a senior investigator; difficulty labels were assigned before any agent evaluation. Each draft problem was further reviewed by an ensemble of LLMs for ambiguous wording, incomplete specifications, and errors, and refined manually. This cycle was repeated until every problem satisfied three criteria: \textit{correctness}, providing sufficient and accurate detail for the simulation to be reproduced; \textit{credibility}, reflecting a realistic molecular simulation research scenario of appropriate complexity; and \textit{clarity}, defining the expected output format and evaluation criteria unambiguously (App.~\ref{app:task_prompt} gives an example). Finally, the trajectory of every reference simulation was inspected visually in VMD or OVITO to confirm physically realistic structural evolution and equilibration.

\textbf{Validation by human experts.} To test whether the reference values can be reproduced independently from the problem descriptions, five evaluators working in MD simulation (four PhD students and one Master's student) each simulated 10 systems, five in \gromacs{} and five in \lammps{}, balanced across difficulty levels, using the same structure and topology files supplied to the agents. \rev{Each reported quantity is compared with its reference value $x_{\mathrm{ref}}$ under the same tolerance used to score agents (Appendix \ref{app:tolerance}). Of the 95 physical quantities evaluated across these 50 systems, all agree with the reference values. This agreement rate also provides a human reference for the agent results in Section~\ref{sec:results}.}

\begin{figure}[ht]
    \centering
    \includegraphics[width=0.9\linewidth]{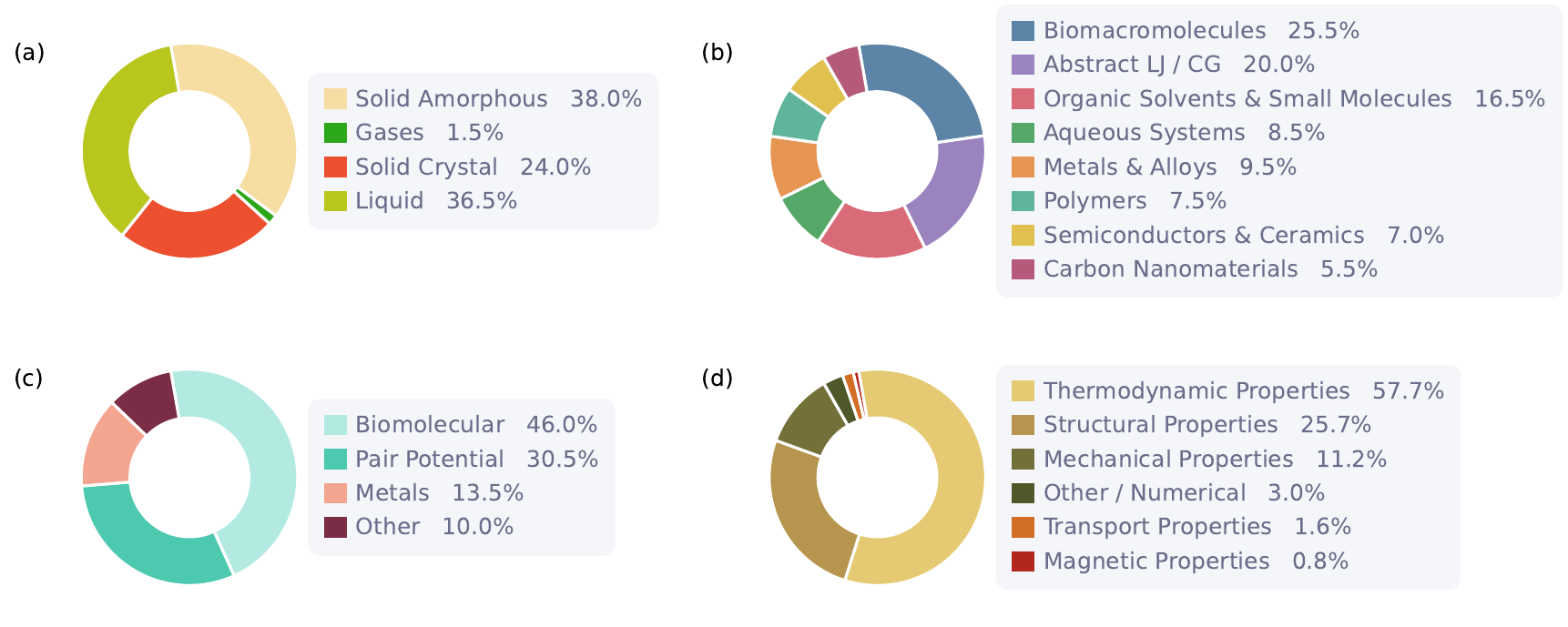}
    \vspace{-0.2in}
    \caption{Statistical overview of the molecular dynamics simulation dataset across four classification dimensions. Donut charts showing the distribution of simulations by (a) physical state, (b) material/system type, (c) force field category, and (d) computed property type.
    }
    \label{fig:data_categories}
    \vspace{-0.1in}
\end{figure}

\textbf{Dataset statistics.} \mdgym{} contains 200 simulation problems with 366 target quantities over 76 distinct properties, together with 160 knowledge questions (Fig.~\ref{fig:data_categories}). \gromacs{} is generally preferred for biomolecular simulations of proteins, lipids, and nucleic acids, whereas \lammps{} offers the flexibility suited to materials, polymers, and coarse-grained systems (App.~\ref{app:engine_diff}); the two engines together cover biomolecules, metals and alloys, polymers, and carbon-based nanomaterials in crystalline and amorphous solid, liquid, and gaseous states. The properties fall into six classes, with the number of distinct properties in parentheses: thermodynamic (24), mechanical (24), structural (13), transport (6), magnetic (2), and other or numerical (7). Thermodynamic properties account for 58\% of target quantities, reflecting their importance in standard MD workflows (App.~\ref{app:property_categories}).

The problems are divided into three levels of difficulty according to the properties they require. \textit{Level~1} (Easy) problems require thermodynamic properties, such as temperature, pressure, and total energy per particle, that need minimal post-processing of the native engine output. \textit{Level~2} (Medium) problems require properties that need substantial post-processing, such as coordination numbers, the average radius of gyration, hydrogen-bond statistics, and energy components other than the potential or total energy; the level is assigned by the property, irrespective of whether the agent computes it during the run or afterwards. \textit{Level~3} (Hard) problems require specialized simulations beyond equilibration runs, such as non-equilibrium deformation, shear, or tensile protocols and multi-stage runs, or post-processing by curve fitting, such as extracting a diffusion coefficient from the slope of the mean-squared displacement; examples include Young's modulus, shear viscosity, diffusion coefficients, thermal conductivity, and magnetic properties.


Since every problem requires running simulations, \mdgym{} is considerably more expensive per problem than language-only or code-generation benchmarks: under the time budget of Sec.~\ref{sec:evaluation}, a single evaluation pass of one agent--model configuration requires about 82 hours of wall-clock time. The size of 200 problems balances coverage of MD workflows against the cost of evaluating many configurations, and Table~\ref{tab:tasks_comparison} places it in the context of existing benchmarks. The problem set can be expanded systematically by varying the material, force field, or cutoff within existing simulation templates, without redesigning the underlying problems.

\textbf{Evaluation harness.} \mdgym{} is implemented as a modular evaluation harness with three independently extensible layers (App.~\ref{app:mdgym}). The \textit{agent layer} provides a common interface to any command-line agent: each implementation runs the agent as a subprocess, streams its actions and outputs to a structured trajectory log, and returns control upon completion or timeout. The \textit{orchestration layer} coordinates a single evaluation episode by building the task prompt, invoking the agent, and passing its reported output to the validator, independently of the agent and engine in use. The \textit{validator layer} compares the reported values with the reference values of each engine-specific task and assigns the score defined in Section~\ref{sec:evaluation}. Each layer is defined by an abstract interface, so that a new agent, a new MD engine such as AMBER, NAMD, or OpenMM, or a new scoring metric is added by implementing a single class without modifying the other components.


\vspace{-0.1in}

\section{Evaluation setup}
\label{sec:evaluation}
\textbf{Agents and models.} We evaluate three general-purpose command-line agent frameworks, Claude Code, Codex, and OpenHands~\citep{wang2024openhands}, each with four LLMs: Claude Opus 4.8, GPT 5.5, DeepSeek V4 Pro~\citep{xu2026deepseek}, and Qwen 3.8 27B~\citep{qwen38}, giving twelve agent--model configurations. All models are accessed through the OpenRouter API\footnote{\url{https://openrouter.ai/}}
, and models with configurable reasoning effort run at the provider's default. Every agent runs headlessly with permission prompts disabled, so that it executes shell commands and edits files autonomously. The system prompts, tool definitions, and configuration of each framework are reported in App.~\ref{app:agent_architectures}. The execution environment provides \lammps{} (22~Jul~2025) and \gromacs{} (v2026.0) together with the required Python packages, and each problem supplies its starting files, such as structure, topology, and potential files. To evaluate the contributions of model and framework further, App.~\ref{app:additional_results} reports additional combinations of eight LLMs and the three frameworks (24 combinations) on a subset of 50 Medium and Hard problems.

\textbf{Task prompt.} Each problem is presented as a single structured prompt generated from its JSON specification (Section~\ref{sec:mdgym}; example in App.~\ref{app:task_prompt}). The prompt states the problem identifier, the target engine, a natural-language description of the simulation, the required output metrics, and the paths to the supplied input files. Agents write all scripts in a dedicated working directory and report their results in a \texttt{final\_answer.json} file that maps each metric name to a unitless numerical value; a problem whose file is missing or cannot be parsed receives a score of zero.

\textbf{Time budget.} Each problem has a time budget of $300 + 3\,t_{\text{sim}}$ seconds, where $t_{\text{sim}}$ is the wall-clock runtime of the reference simulation measured during ground-truth generation on 24-core AMD CPUs. The fixed 300~seconds allows the agent to read the task and plan, and the factor of 3 allows about three complete simulation attempts; an agent that exceeds the budget is terminated and the problem is scored as failed. We express the budget in wall-clock time since simulation execution dominates the cost of each problem, and token budgets are not comparable across models with different tokenizers. 

\textbf{Evaluation metrics.} For a problem with required metric set $M$, the score is $s = \frac{1}{|M|} \sum_{k \in M} \mathbf{1}\left[\text{passed}(k)\right]$,
where $\text{passed}(k) = 1$ if the reported value of metric $k$ lies within the tolerance defined in App.~\ref{app:tolerance}, and 0 otherwise. A problem is solved only if $s = 1$. For each configuration, we report the \textit{full-success rate}, the fraction of problems solved, and the \textit{total partial-credit score}, the sum of $s$ over problems, each by difficulty level and engine together with a 95\% Wilson confidence interval (App.~\ref{appendix:stats}) over the same set of problems for every configuration. In addition, we inspect trajectories manually for answers that were guessed or recalled without running the simulation, and score such problems as failed. \rev{For the knowledge questions and MD-QA, each question is multiple choice, having only one correct option. The model selects one option, and we report \textit{accuracy}, the fraction of questions for which the selected option matches the answer key.}


\textbf{Failure analysis.} Three complementary analyses characterize how agents fail (App.~\ref{app:error_modes}). A rule-based classifier assigns each failed trajectory to categories of an expert-curated taxonomy. An LLM judge (Claude Sonnet 5, majority vote over three classifications) assigns each engine error to predefined \lammps{} and \gromacs{} categories. A second LLM judge (Claude Sonnet 5) determines the last pipeline stage that each simulation completed. To isolate post-processing, we additionally provide agents with correct simulation outputs for 50 problems and ask them only to compute the target observables. To determine whether failures arise from missing domain knowledge, we evaluate the same models on MD-QA~\citep{rios2026ai} and on the \mdgym{} knowledge questions, a curated set of questions which require expert-level knowledge in MD simulations to answer.

\section{Results and Discussion}
\label{sec:results}

\begin{figure}[ht]
    \centering
    \includegraphics[width=0.9\linewidth]{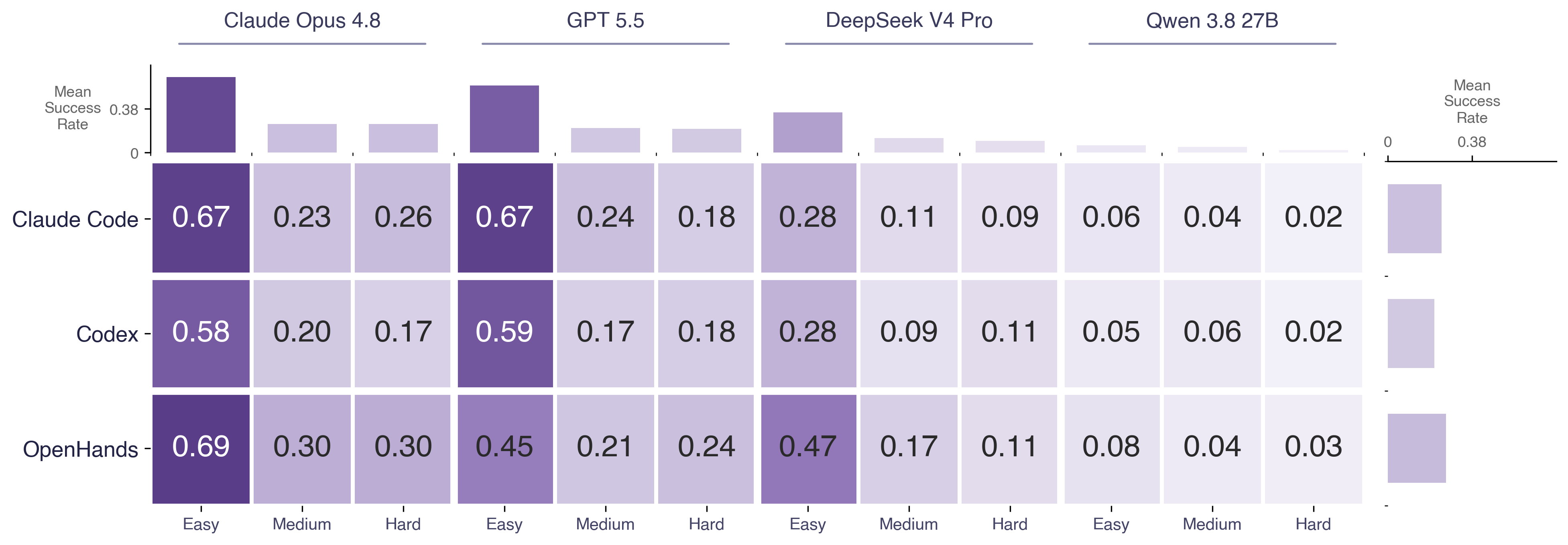}
    \vspace{-0.2in}
    \caption{Full-success rate of each agent--model configuration by difficulty level on all \mdgym{} tasks. Bars above and to the right give mean values over frameworks and models, respectively.}
    \vspace{-0.2in}
    \label{fig:combined_success_rate}
\end{figure}

\begin{figure}[ht]
    \centering
    \includegraphics[width=0.9\linewidth]{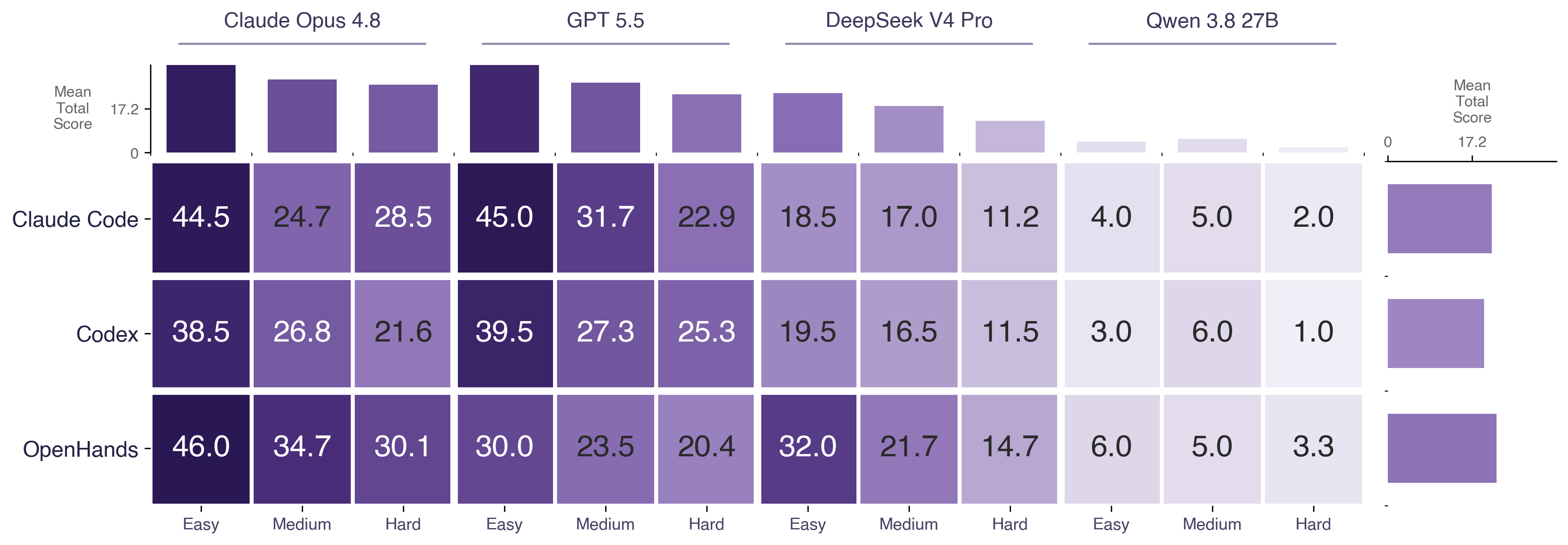}
    \vspace{-0.1in}
    \caption{Partial-credit score, the sum of per-problem scores $s$, of each agent--model configuration by difficulty level. Bars above and to the right give means over frameworks and models, respectively.}
    \vspace{-0.1in}
    \label{fig:combined_total_score}
\end{figure}

\paragraph{Overall Performance.}
Figures~\ref{fig:combined_success_rate} and~\ref{fig:combined_total_score} report the full-success rate and the total partial-credit score of the twelve agent--model configurations. On Easy problems, the best configuration of each framework solves \rev{67}\% (Claude Code with Claude Opus 4.8 or GPT 5.5), 59\% (Codex with GPT 5.5), and \rev{69}\% (OpenHands with Claude Opus 4.8) of the problems. These rates fall to 24\%, 20\%, and 30\% on Medium problems and to 26\%, \rev{18}\%, and 30\% on Hard problems. OpenHands with Claude Opus 4.8 attains the highest rate at every difficulty level, with a 95\% Wilson interval of \rev{56.6--78.8}\% on Easy problems (App.~\ref{appendix:stats}). The two proprietary models outperform DeepSeek V4 Pro and Qwen 3.8 27B under every framework and at every difficulty level, with one exception: GPT 5.5 under OpenHands on Easy problems (\rev{45\% against 47\%} for DeepSeek V4 Pro). Pooled over frameworks, the mean per-problem scores are \rev{0.492, 0.443, 0.271, and 0.059} for Claude Opus 4.8, GPT 5.5, DeepSeek V4 Pro, and Qwen 3.8 27B. All pairwise differences between models are significant (Mann--Whitney U, $p < 0.001$) except that between Claude Opus 4.8 and GPT 5.5 \rev{($p = 0.052$)}. App.~\ref{app:additional_results} reports the remaining model--framework combinations. Doubling the time budget from 300 to 600~seconds on a subset of 50 problems improves the performance of Claude Sonnet 4.6 by only 4\%. \rev{Repeated runs of two configurations on a 50-problem subset confirm that these results are stable: success rates have a standard deviation of 1.2–2.0 percentage points across runs, and the ranking of models is the same in every run (App.~\ref{appendix:repeats}).}

Disaggregated by engine (Figs.~\ref{fig:lammps_success_rate} and~\ref{fig:gromacs_success_rate}), ten of the twelve configurations achieve a higher Easy success rate on \gromacs{} than on \lammps{}. The best Easy rate reaches 93\% on \gromacs{} (GPT 5.5 under Claude Code) against 53\% on \lammps{} (Claude Opus 4.8 and GPT 5.5 under OpenHands). This gap narrows with difficulty. \rev{On Hard problems, Claude Opus 4.8 solves a higher fraction of \lammps{} problems than of \gromacs{} problems under two of three frameworks: 33\% against 17\% under Claude Code, 14\% against 20\% under Codex, and 39\% against 20\% under OpenHands.} From Easy to Hard, the success rates of Claude Opus 4.8 and GPT 5.5 decrease by 57 to 73 percentage points on \gromacs{} and by 14 to 33 percentage points on \lammps{}. The one exception on \gromacs{} is GPT 5.5 under OpenHands, which decreases by 27 points. The decrease in per-problem score with difficulty is significant for Claude Opus 4.8, GPT 5.5, and DeepSeek V4 Pro (Kruskal--Wallis, $p \le 0.003$) and across all twelve configurations (\rev{$H = 52.20$}, $p < 0.001$). Qwen 3.8 27B, whose scores are already close to zero on Easy problems, shows a trend in the same direction that is not significant (\rev{$p = 0.051$}).

\begin{wrapfigure}{r}{0.7\textwidth}
        \centering
        \vspace{-0.15in}
        \includegraphics[width=\linewidth]{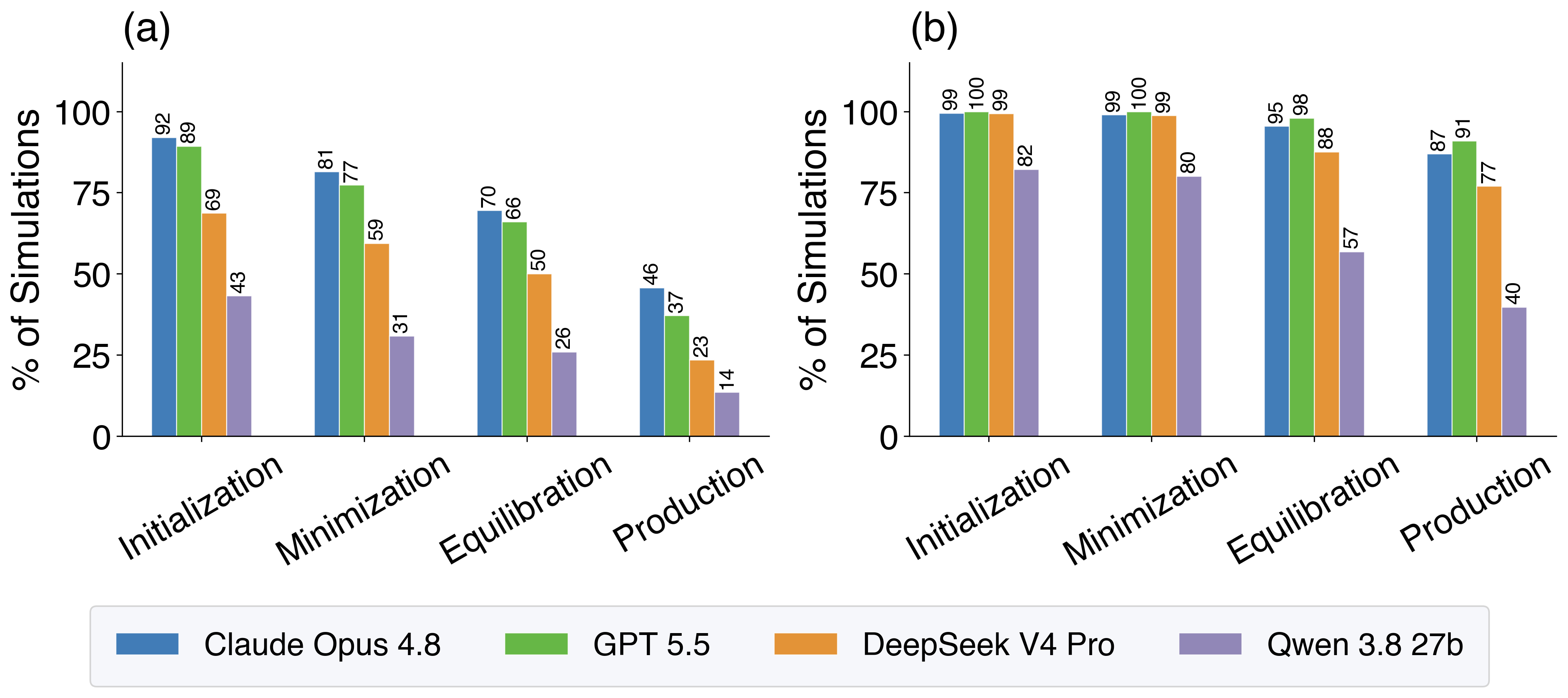}
        \vspace{-0.15in}
        \caption{Percentage of runs completing each stage of the MD pipeline for each model, sum pooled over all harnesses, for (a) \lammps{} and (b) \gromacs{} problems. Stages are assigned by the classifier of App.~\ref{app:error_modes}.}
        \label{fig:simulation_completion_funnel}
        \vspace{-0.15in}
\end{wrapfigure}

Varying the framework at fixed model changes the success rate far less than varying the model at fixed framework. On Easy problems, Claude Opus 4.8 solves \rev{67.2\%, 57.8\%, and 68.8\%} of the problems under Claude Code, Codex, and OpenHands, and the three 95\% Wilson intervals overlap. Under OpenHands, the Easy success rate ranges from 8\% for Qwen 3.8 27B to 69\% for Claude Opus 4.8. A two-way analysis of variance on per-problem scores confirms this asymmetry. Model identity explains \rev{16.4}\% of the variance (\rev{$F = 158.27$}, $p < 0.001$), framework identity explains 0.14\% and is not significant (\rev{$F = 2.06$}, $p = 0.127$), and their interaction explains 0.98\% ({$F = 4.74$}, $p < 0.001$). The controlled comparison of eight LLMs under all three frameworks on 50 Medium and Hard problems gives the same result (Fig.~\ref{fig:all_combinations}). For a fixed proprietary model, the frameworks differ by at most three problems, while across models the success rate ranges from 0\% to 16\%. The base model thus determines performance on \mdgym{} far more than the agent framework does.



\textbf{Knowledge and execution.} The same models answer MD knowledge questions accurately. On MD-QA, Claude Opus 4.8, GPT 5.5, DeepSeek V4 Pro, and Qwen 3.8 27B score 90\%, 95\%, 97.5\%, and 88.75\%, and on the 160 \mdgym{} knowledge questions they score 90\%, 90\%, 88\%, and 80\%. The contrast with the end-to-end success rates places the difficulty in applying this knowledge to configure, execute, and analyze a simulation, which the following research questions examine.

\textbf{\textit{RQ1: Where in the simulation pipeline do agents fail?}}
Figure~\ref{fig:simulation_completion_funnel} reports the percentage of runs that complete each stage of the pipeline. Attrition is far steeper on \lammps{} than on \gromacs{}. Claude Opus 4.8 declines from 92\% of runs at initialization to 46\% at production on \lammps{}, and from 99\% to 87\% on \gromacs{}, consistent with the engine-level gap in success rates. On \lammps{}, Claude Opus 4.8, GPT 5.5, and DeepSeek V4 Pro lose the largest share of runs, 24 to 29\%, between equilibration and production. Model capability shapes both the level and the shape of the curve. Claude Opus 4.8 and GPT 5.5 track each other closely on both engines, DeepSeek V4 Pro trails by a margin that widens at each stage. Qwen 3.8 27B fails earliest; 57\% of its \lammps{} runs and 18\% of its \gromacs{} runs do not complete initialization, and only 14\% and 40\% reach production. That said, completing production does not ensure a correct result. Claude Opus 4.8 and GPT 5.5 complete production in 87\% and 91\% of their \gromacs{} runs, and the remaining failures lie in the reported values, which RQ2 examines.

\begin{wrapfigure}{r}{0.6\textwidth}
    \centering
    \vspace{-0.1in}
    \includegraphics[width=0.48\textwidth]{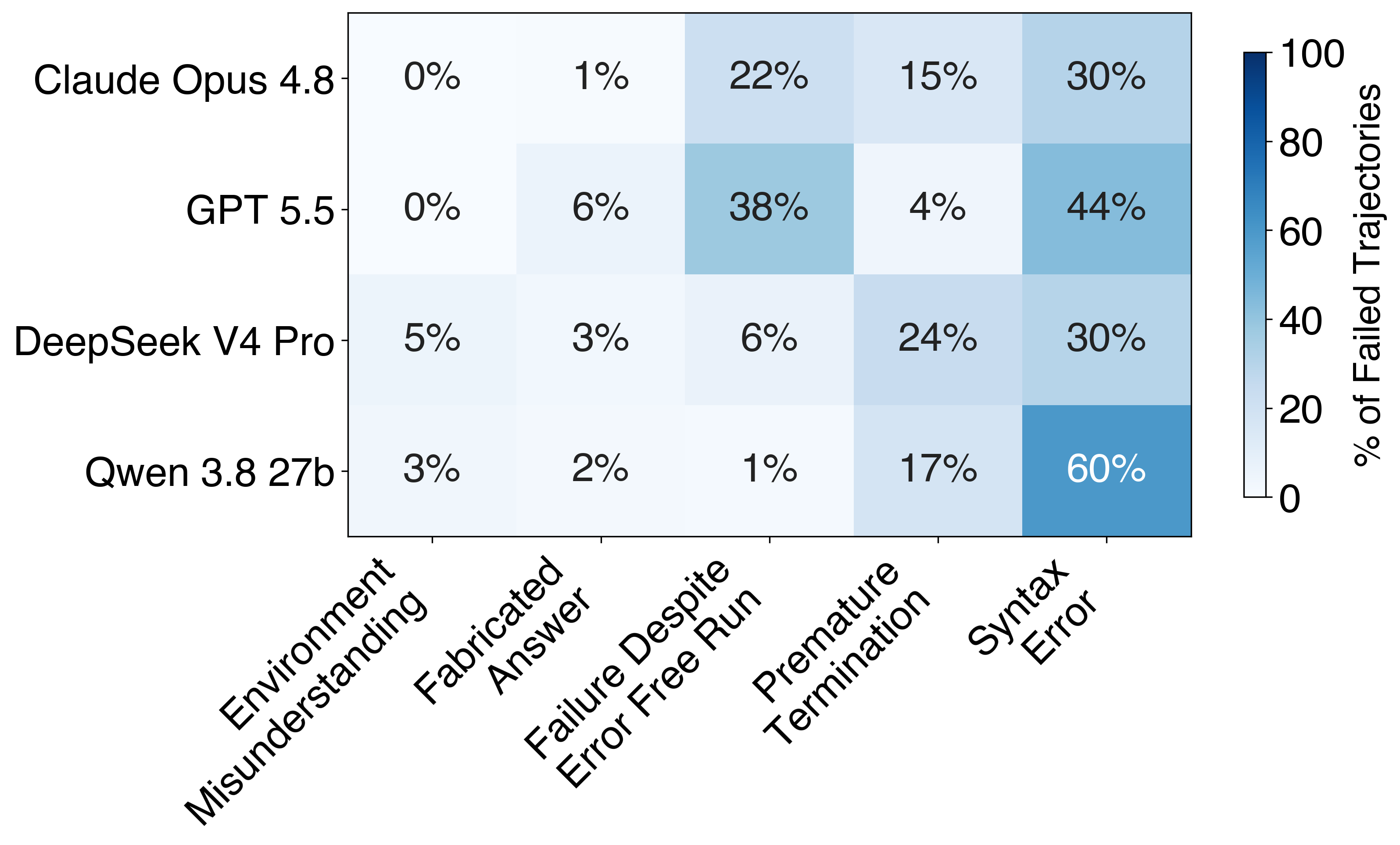}
    \vspace{-0.1in}
    \caption{\rev{Prevalence of failure modes, as the percentage of failed trajectories exhibiting each mode. Modes are non-exclusive.}}
    \label{fig:heatmap}
    \vspace{-0.1in}
\end{wrapfigure}

\textbf{\textit{RQ2: What are the most common failure modes?}} 
Figure~\ref{fig:heatmap} reports the prevalence of five failure modes among failed trajectories (App.~\ref{app:error_modes}). Syntax errors are the most prevalent mode for every model (30--60\%), whereas fabricated answers (at most 6\%) and environment misunderstanding (at most 5\%) are rare. When agents fabricate, they either recall a plausible value from parametric memory or approximate it using standalone Python code instead of the simulation. Interestingly, Claude Opus 4.8 and DeepSeek V4 Pro both encounter syntax errors in 30\% of their failed trajectories, yet Claude Opus 4.8 achieves a higher success rate across all frameworks and at every difficulty level. Claude Opus 4.8 and DeepSeek V4 Pro recover 18\% and 14.3\% of the time after encountering a syntax error.
The remaining modes separate the models. Premature termination is most frequent for DeepSeek V4 Pro (24\%) and Qwen 3.8 27B (17\%), and least frequent for GPT 5.5 (4\%). \rev{Failure Despite Error-Free Run, which comprises every failed trajectory whose answer came from an error-free run, accounts for 22\% and 38\% of the failures of Claude Opus 4.8 and GPT 5.5, against 6\% and 1\% for DeepSeek V4 Pro and Qwen 3.8 27B. Weaker models therefore fail early in the pipeline, and stronger models, having cleared the early stages, fail to produce a physically correct simulation or obtain an incorrect value from a completed simulation.} The drop from produced to correct answers at every difficulty level shows the same pattern (Fig.~\ref{fig:task_completion_funnel}).

\par

Failure Despite Error-Free Run combines two sources of error: the analysis of a correct simulation and simulations that completed without error but were physically incorrect (Table~\ref{tab:failure_categories}). To isolate the first, we perform a targeted ablation. Given correct simulation outputs for 50 problems and asked only to compute the target observables, Claude Sonnet 4.6 and DeepSeek V4 Pro each succeed on 72\% of the problems, against 20\% and 44\% end-to-end on the same problems. Post-processing is therefore an independent source of error, and the remaining failures in this category reflect simulations that were configured incorrectly without raising any error.

\textbf{\textit{RQ3: Do agents fail at syntax, specification, or physics?}} We group the engine errors of Figure~\ref{fig:engine_specific_errors} into four classes: \textit{syntax} errors, where the engine cannot parse a command; \textit{specification} errors, where individually valid commands define an incomplete or inconsistent simulation through its force-field assignments, data and topology files, atom groups, or command order; \textit{execution-environment} errors, where the input requests parallel or accelerator features that the CPU-only environment does not provide; and \textit{runtime instabilities}, where a running simulation becomes numerically or physically unstable. In \lammps{}, syntax errors (Command Syntax and Variable/Expression) account for 28.2\% of all errors and specification errors for 47.0\%, led by Pair/Force Field mis-specification (13.1\%) and Data File and Read errors (10.6\%). Execution-environment errors (Package/Partition and MPI/Parallel) account for 14.0\%, and runtime instabilities, led by Energy Divergence (3.9\%) and Lost Atoms (3.5\%), for 10.7\%. In \gromacs{}, 74.3\% of errors arise during preprocessing by \texttt{grompp}, before the first MD step, and are dominated by Index Group (47.2\%) and Topology (16.6\%) errors. Of the remaining 25.7\%, runtime instabilities (Pressure Coupling, Energy Blow-up etc.) account for 10.0\%, and errors in electrostatics, box, and output settings (PME/Electrostatics, PBC/Box, and Output/File Write) for 15.0\%.

In both engines, most engine errors therefore arise while agents assemble the simulation input, and a minority arise from the instability of a running simulation. The same models answer 88--90\% of the \mdgym{} knowledge questions correctly, and 80\% in the case of Qwen 3.8 27B. They know the individual commands and concepts but fail to combine interdependent choices of force field, topology, atom groups, and execution settings into a specification consistent with the target engine and hardware. Physically incorrect simulations that run to completion raise no engine error; they appear in RQ2 among the incorrect answers from error-free runs.

\textbf{\textit{RQ4: What do these failures collectively reveal?}} The analyses identify three capabilities that current agents lack for MD simulation. The first is \textit{coherent specification}: combining the structure, force field, topology, atom groups, and execution settings into an input consistent with the target engine, the capability whose absence produces most engine errors in RQ3. The second is \textit{domain-aware error recovery}: interpreting error messages and correcting the input accordingly, whose absence appears as premature termination in 4--24\% of failed trajectories and as the stage attrition in RQ1. The third is \textit{physical and methodological validation}: recognizing when a completed simulation or its analysis yields a physically incorrect value, and selecting the analysis procedure appropriate to the measured quantity; its absence accounts for 22\% and 38\% of the failures of Claude Opus 4.8 and GPT 5.5, and for the 28\% of problems that remain unsolved when agents receive correct simulation outputs. 

Each of these gaps guides future intervention strategies. The engine error logs collected in \mdgym{} provide pairs of error messages and corrective actions for fine-tuning recovery behavior. \lammps{} input scripts and \gromacs{} parameter and topology files are low-resource languages, underrepresented in pretraining corpora relative to Python or C++, which motivates targeted data collection and continued pretraining on simulation inputs. Physics-based validators and engine-specific error parsers can be added to the agent loop through the modular harness of \mdgym{}. The same requirements arise in other scientific computing domains with multi-stage pipelines and physically defined correctness criteria, such as density functional theory, finite element analysis, and computational fluid dynamics, where software engineering benchmarks likewise do not measure them.


\vspace{-0.1in}
\section{Conclusion}
\vspace{-0.05in}
\label{sec:limitations}
We introduce \mdgym{}, a benchmark of 200 expert-curated MD simulation problems in \lammps{} and \gromacs{} at three difficulty levels, with 366 target quantities scored against reference values from expert simulations. Across twelve configurations of three general-purpose command-line agent frameworks and four LLMs, the best configuration solves \rev{69}\% of Easy problems, \rev{30}\% of Medium problems, and \rev{30}\% of Hard problems. The base model determines performance far more than the framework. At fixed model, the Easy success rate varies by 3 to 22 percentage points across frameworks, whereas at fixed framework it varies by up to 61 percentage points across models. The same models answer 80--90\% of the \mdgym{} knowledge questions correctly, which locates the difficulty in applying domain knowledge to configure, execute, and analyze a simulation. Error analysis reveals that most engine errors arise while agents assemble the simulation input, and few from the instability of a running simulation. Weaker models fail early in the pipeline, through syntax errors and premature termination, whereas stronger models complete more simulations but report incorrect values from error-free runs in \rev{22--38\%} of their failed trajectories. These results identify three capabilities that agents must acquire to serve as reliable partners in computational science: coherent identification and specification of simulation inputs, domain-aware recovery from engine errors, and physical and methodological validation of results. We release the benchmark, the evaluation harness, and the agent trajectories and will maintain a public leaderboard to track progress on these capabilities. \mdgym{} code and dataset are available at \href{https://anonymous.4open.science/r/MDBench-C1F5}{code} and \href{https://anonymous.4open.science/r/MDGym-Dataset-CFE8}{dataset}.


\textbf{Limitations and future work.} \mdgym{} can be extended to include additional MD engines such as AMBER, NAMD, or OpenMM. All problems use classical force fields and exclude ab initio MD and enhanced sampling methods such as metadynamics and replica exchange, which require additional methodological choices and form a natural extension to harder problems. We evaluate general-purpose command-line frameworks, some of whose system prompts carry software-engineering framing (App.~\ref{app:agent_architectures}); development of agents designed for simulation, with physics-based validators or engine-specific repair loops, could further improve the performance.  \rev{Because of the high cost of the runs ($\sim$4{,}000 USD for the entire project), each configuration was evaluated once on the full benchmark; repeated runs of two configurations on a 50-problem subset show stable success rates and a consistent ranking of models (App.~\ref{appendix:repeats}). Extending them to all configurations is left to future work. Finally, extending the benchmark to open-ended problems, in which agents must choose the method, judge convergence, and report the uncertainty of their results, would evaluate the scientific judgment that autonomous discovery requires.}


\subsection*{AI use statement}

In this work, we used generative AI tools for code generation and text improvement. We have reviewed all AI-assisted work. We take responsibility for the final content of this work,
including text, claims or artefacts produced with the aid of generative AI.




\subsection*{Reproducibility statement}
The complete \mdgym{} benchmark (\url{https://anonymous.4open.science/r/MDGym-Dataset-CFE8}), including all 200 tasks, their JSON specifications (problem descriptions, required metrics, ground-truth values, and starting files), and the exact task prompts sent to agents, is available at the anonymized repository. Our evaluation harness code (\url{https://anonymous.4open.science/r/MDBench-C1F5}), including the agent, orchestration, and validator layers described in Appendix~\ref{app:mdgym}, is publicly released. The exact prompts and classification rules used for our automated trajectory-level and simulation-level error analyses are provided in full in Appendix~\ref{app:judge_prompt}, including the complete system prompts given to the LLM-based error classifier. Our statistical methodology, including the exact tests and software packages used to compute all reported significance results, is described in Appendix~\ref{appendix:stats}, and the raw per-problem scores underlying these analyses are released alongside the benchmark to enable independent verification. Details of our human expert validation study, including the number of evaluators, their qualifications, and the exact validation protocol, are provided in Section~\ref{sec:mdgym}.

\bibliography{iclr2027_conference}
\bibliographystyle{iclr2027_conference}

\newpage
\appendix
\renewcommand{\thefigure}{\thesection.\arabic{figure}}
\setcounter{figure}{0}
\renewcommand{\thetable}{\thesection.\arabic{table}}
\setcounter{table}{0}

\section{Molecular dynamics simulation}
\label{app:md_protocol}

This appendix summarizes the physical and numerical requirements of MD simulation that \mdgym{} tests, describes the stages of the simulation pipeline used, and relates each requirement to the engine error categories of App.~\ref{app:error_modes}.

\subsection{Equations of motion and numerical integration}
MD simulation computes the time evolution of a system of $N$ interacting atoms by numerically integrating the classical equations of motion. Given positions $\{\mathbf{q}_i\}_{i=1}^N$ and momenta $\{\mathbf{p}_i\}_{i=1}^N$, the dynamics follow Hamilton's equations,
\begin{equation}
\frac{d\mathbf{q}_i}{dt} = \frac{\partial H}{\partial \mathbf{p}_i}, \qquad
\frac{d\mathbf{p}_i}{dt} = -\frac{\partial H}{\partial \mathbf{q}_i},
\end{equation}
where $H = \sum_i \mathbf{p}_i^2/2m_i + V(\{\mathbf{q}_i\})$ is the Hamiltonian, $m_i$ are the atomic masses, and $V$ is the potential energy surface. These equations are equivalent to Newton's equations of motion, with the force on each atom given by $\mathbf{F}_i = -\partial V/\partial \mathbf{q}_i$. The forces are evaluated at every step, and their evaluation dominates the computational cost of a simulation.

Integration is almost universally performed with the velocity Verlet algorithm~\citep{verlet1967computer}, which advances the system by a timestep $\Delta t$ as
\begin{align}
\mathbf{p}_i\!\left(t+\tfrac{\Delta t}{2}\right) &= \mathbf{p}_i(t) + \tfrac{\Delta t}{2}\,\mathbf{F}_i(t), \nonumber\\
\mathbf{q}_i(t+\Delta t) &= \mathbf{q}_i(t) + \tfrac{\Delta t}{m_i}\,\mathbf{p}_i\!\left(t+\tfrac{\Delta t}{2}\right), \\
\mathbf{p}_i(t+\Delta t) &= \mathbf{p}_i\!\left(t+\tfrac{\Delta t}{2}\right) + \tfrac{\Delta t}{2}\,\mathbf{F}_i(t+\Delta t), \nonumber
\end{align}
where the second half-step in momentum uses the forces evaluated at the updated positions. The algorithm is symplectic and time-reversible. It therefore conserves a shadow Hamiltonian close to the true Hamiltonian and exhibits no long-term energy drift, provided $\Delta t$ is sufficiently small. Without a thermostat or barostat, the integration samples the microcanonical (NVE) ensemble, in which the number of particles $N$, the volume $V$, the total energy $E$, and the total linear momentum are conserved. The drift of these quantities over a run is a standard diagnostic of integration quality.

The timestep, typically of the order of a femtosecond, must be a small fraction of the period of the fastest vibrational mode in the system. In atomistic biomolecular simulations, the O--H stretch near 3500~cm$^{-1}$ has a period of about 10~fs and limits the timestep to $\Delta t \lesssim 1$~fs, unless bond constraints are imposed through algorithms such as SHAKE, RATTLE, or LINCS. When the timestep is too large, the shadow Hamiltonian departs from the true Hamiltonian and the integration becomes unstable, with the energy drifting without bound. Marginally unstable choices may not diverge visibly, but they sample incorrect regions of phase space. In \mdgym{}, such failures appear as the Timestep, Energy Divergence, Lost Atoms, and Constraint Failure categories in \lammps{}, and as the LINCS/Constraint and Energy Blow-up categories in \gromacs{} (App.~\ref{app:error_modes}).

\subsection{Force Fields}

The empirical potential $V(\{\mathbf{q}_i\})$ approximates the true quantum mechanical potential energy surface as a sum of bonded and non-bonded terms. For molecular systems, the standard decomposition is

\begin{align}
    V = &\sum_{\text{bonds}} k_b(r - r_0)^2 
  + \sum_{\text{angles}} k_\theta(\theta - \theta_0)^2
  + \sum_{\text{dihedrals}} \sum_n \frac{V_n}{2}[1 + \cos(n\phi - \delta)]\\
  & + \sum_{i<j} \left[ 4\varepsilon_{ij}\left(\frac{\sigma_{ij}^{12}}{r_{ij}^{12}} 
    - \frac{\sigma_{ij}^6}{r_{ij}^6}\right) 
    + \frac{q_i q_j}{4\pi\varepsilon_0 r_{ij}} \right],
\end{align}

where the parameters $\{k_b, r_0, k_\theta, \theta_0, V_n, \varepsilon_{ij}, 
\sigma_{ij}, q_i\}$ constitute the force field constants that are optimised to reproduce target properties either obtained from experiments or from first principle simulations. For materials systems, simpler pair potentials (Lennard-Jones, Morse) or many-body potentials (EAM for metals, Tersoff and AIREBO for covalent systems, ReaxFF for reactive chemistry) replace the molecular mechanics decomposition. Each force field is parameterized against a specific class of systems and a specific range of thermodynamic conditions. Applying a force field outside its parameterization domain, mixing parameters from incompatible force fields, or neglecting long-range electrostatics when partial charges are significant, all produce forces that bear no meaningful relationship to the true interatomic interactions. The engine accepts these inputs without any error and integrates them providing no feedback to the user about the correctness.

\subsection{Thermodynamic Ensembles and Equilibration}

The basic velocity Verlet integrator samples the NVE ensemble. Most physical measurements require NVT or NPT conditions, achieved by coupling the system to a thermostat (V-rescale, Nos\'{e}-Hoover) or barostat (Parrinello-Rahman, Berendsen) with associated coupling time constants $\tau_T$ and $\tau_P$. These must be matched to the physical relaxation timescales of the system. Aggressive coupling suppresses the fluctuations that define the ensemble and introduces artifacts in dynamical properties; weak coupling prevents equilibration within any practical simulation window. Equilibration itself proceeds in stages: energy minimization via steepest descent to remove steric clashes, followed by NVT and NPT runs of sufficient length that the relevant observables, such as potential energy, density, and pressure, have converged to stable mean values. Insufficient equilibration produces trajectories that are thermodynamically inconsistent with the intended state, without any engine-level indication that this has occurred.

\subsection{Physical Validity Conditions}

Beyond integration stability, a physically valid simulation must satisfy several conditions that the engine does not enforce. \textbf{System size} must be large enough that periodic images do not introduce spurious correlations; properties sensitive to long-wavelength fluctuations, such as elastic constants, transport coefficients, interfacial tensions, converge to their thermodynamic limit only when the box dimension substantially exceeds the relevant correlation length. For non-equilibrium protocols, such as uniaxial straining for elastic moduli, planar shear for viscosity, imposed thermal gradients for conductivity, the applied perturbation must lie within the \textbf{linear response regime}; at high loading or shear rates the measured response is rate-dependent and does not reflect an intrinsic material property. 
\textbf{Post-processing} introduces its own requirements: computing a diffusion coefficient from mean-squared displacement requires identifying the Fickian regime and fitting over an appropriate time window; extracting a glass transition temperature requires fitting two linear segments to the density-temperature curve and locating their intersection; computing a viscosity via the Green-Kubo relation requires integrating a stress autocorrelation function to convergence. The same trajectory yields different answers depending on how the analysis is performed, and the engine provides no guidance on which analysis is physically correct.

Collectively, these requirements---timestep stability, force field validity, ensemble consistency, finite-size convergence, linear response, and methodological precision in post-processing---must be satisfied simultaneously. None is enforced automatically. A simulation that violates any one of them runs to completion, produces output of the expected format, and gives no signal that anything is wrong. This is the regime that separates MD simulation from software engineering tasks, and the regime that \mdgym{} is designed to measure.

\section{AI Agents for Scientific Workflows}
\label{app:ai_agents}
Early systems based on LLMs demonstrated that agents could autonomously plan and execute chemical synthesis. For instance, ChemCrow~\citep{bran_2023chemcrow} integrated 18 expert-designed tools with GPT 4 to automate tasks across organic synthesis, drug discovery, and materials design, while Coscientist~\citep{Boiko_2023} demonstrated autonomous chemical research in a laboratory setting. More recent systems have extended this vision further along the discovery pipeline: The AI Scientist~\citep{lu2024ai} demonstrated fully automated hypothesis generation, experimentation, and paper writing, and its successor~\citep{yamada2025ai} produced manuscripts accepted at a peer-reviewed workshop. In the materials domain, SciToolAgent~\citep{ding_2025scitoolagent} automates hundreds of scientific tools across biology, chemistry, and materials science via a knowledge-graph-driven retrieval system. 

\section{\mdgym: Technical Details}
\label{app:mdgym}
\subsection{Architecture}

\begin{figure}[ht]
    \centering
    \includegraphics[width=1.0\linewidth]{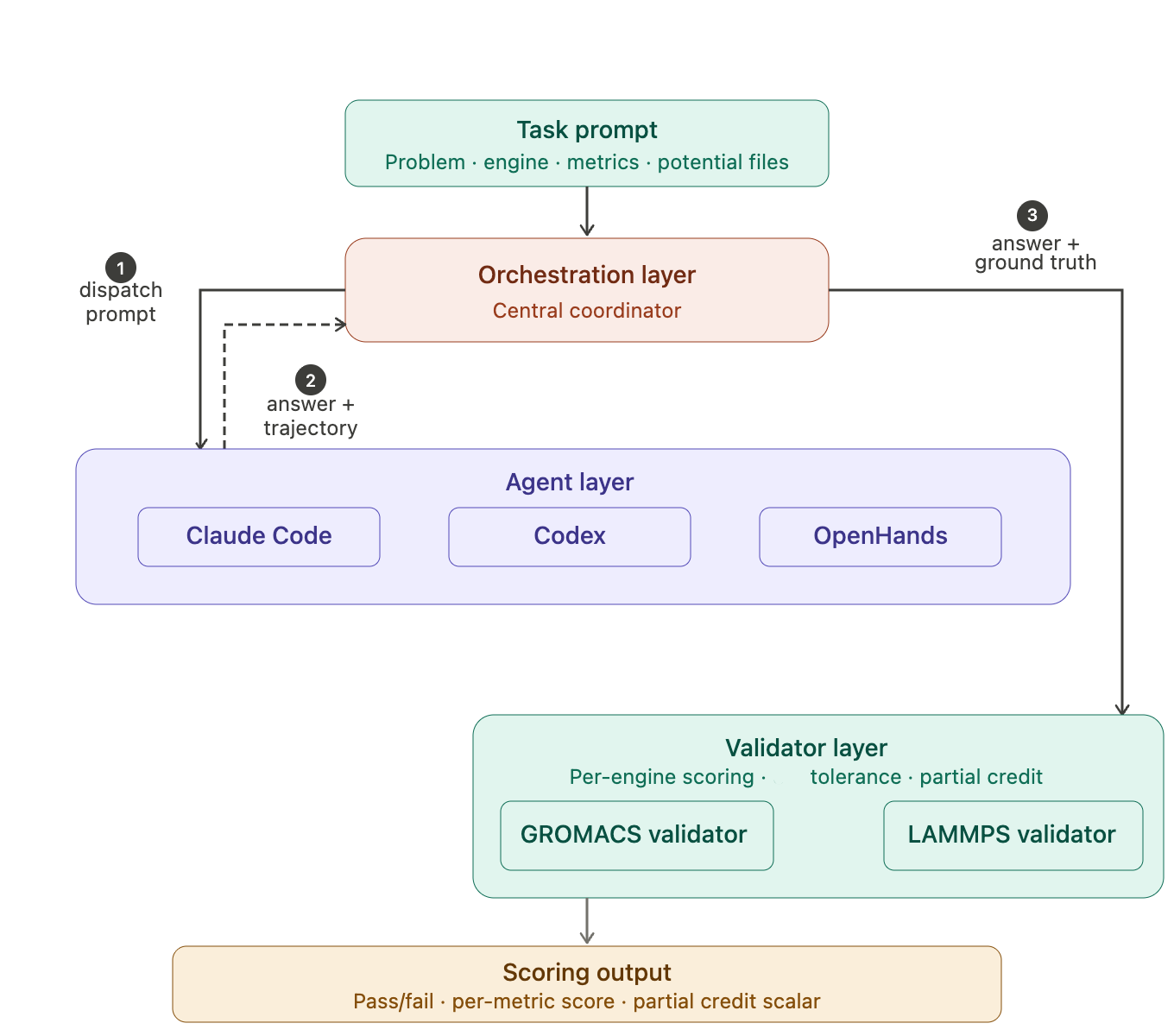}
    \caption{Architecture of the \mdgym{} evaluation framework, showing the three independently extensible layers — agent, orchestration, and validator — and the data flow between them.}
    \label{fig:MDGYM}
\end{figure}

Figure~\ref{fig:MDGYM} illustrates the three-layer architecture of the \mdgym{} evaluation harness. The orchestration layer acts as the central coordinator, dispatching task prompts to the agent layer (Step 1), receiving agent outputs (Step 2), and routing them to the validator layer for automated scoring (Step 3). Each layer is governed by a well-defined abstract interface, making it independently extensible.

\subsubsection{Agent Layer}

All agents implement a common \texttt{BaseAgent} abstract interface that exposes a single \texttt{execute(prompt, working\_dir, timeout, engine)} method. Each concrete implementation wraps the corresponding CLI tool as a subprocess, streams output to a \texttt{trajectory\_log.jsonl} file in the problem's working directory, and returns control to the orchestrator upon completion or timeout. Currently supported agents and their corresponding implementations are: \texttt{ClaudeCodeAgent}, \texttt{CodexAgent}, \texttt{OpenHandsAgent}. Registering a new agent requires only subclassing \texttt{BaseAgent}, implementing the \texttt{execute} method, and adding a single entry to the agent registry via \texttt{AgentFactory.register\_agent(AgentType.NEW\_AGENT, NewAgentClass)}.

\subsubsection{Orchestration Layer}

The \texttt{SimulationOrchestrator} coordinates a single evaluation episode. Its responsibilities are:
\begin{enumerate}
    \item Build the structured task prompt from the problem JSON specification
    \item Invoke the agent via the \texttt{BaseAgent} interface
    \item Read the agent's \texttt{final\_answer.json} output from the working directory
    \item Pass the output and ground truth to the appropriate validator
\end{enumerate}

The orchestrator is constructed via an \texttt{OrchestratorBuilder} that takes an agent type and an engine type as the only required inputs, resolving the concrete agent and validator implementations internally via their respective factories. This keeps the top-level evaluation loop entirely decoupled from the details of any specific agent or engine.

\subsubsection{Validator Layer}

Each MD engine has a corresponding \texttt{BaseValidator} subclass that implements a \texttt{validate(output, working\_dir, ground\_truth)} method and returns a \texttt{ValidationResult} containing:

\begin{itemize}
    \item A binary pass/fail flag
    \item A per-metric breakdown of which quantities passed or failed
    \item A scalar score in $[0, 1]$ representing the fraction of required metrics within relative tolerance of the ground-truth value
\end{itemize}

Currently registered validators are \texttt{LAMMPSValidator} and \texttt{GROMACSValidator}. Adding support for a new MD engine requires subclassing \texttt{BaseValidator} and a single registration call: \texttt{ValidatorFactory.register\_validator(MDEngine.AMBER, AmberValidator)}.

The validator is free to incorporate engine-specific logic, such as reading auxiliary output files, applying engine-specific unit conversions, or using tighter tolerances for specific property categories, without affecting any other part of the framework.

\subsubsection{Session Management and Logging}

In batch mode, the framework processes all problems in a session sequentially, writing per-problem results to isolated subdirectories under a session root. Each trajectory log (\texttt{trajectory\_log.jsonl}) records all agent actions, tool calls, and outputs at the granularity of individual steps, and is automatically converted to a structured JSON format after each problem to facilitate offline analysis of agent reasoning and tool-use patterns. A configurable inter-problem delay mitigates API rate limiting when evaluating cloud-hosted agents. Session-level results are accumulated in a \texttt{results.csv} summary file containing per-problem scores, pass/fail flags, and metadata for downstream analysis.


\subsection{JSON format}
\label{app:json}
The following presents the format of the JSON file in which the tasks are prepared which is passed as input during evaluation. 

\begin{center}
    \begin{tabular}{ll}
        \toprule
        \textbf{Field} & \textbf{Description} \\
        \midrule
        \texttt{id} & Unique problem identifier \\
        \texttt{problem\_description} & Full natural language description of the simulation task \\
        \texttt{level} & 1 (Easy) / 2 (Medium) / 3 (Hard) \\
        \texttt{engine} & {\gromacs} or {\lammps} \\
        \texttt{metrics} & List of physical observables to compute  \\
        \texttt{ground\_truth} & Expected values for each metric $ \pm $ estimated error \\
        \texttt{time\_limit} & Maximum allowed simulation time in seconds \\
        \bottomrule
    \end{tabular}
    \end{center}

\subsection{Tolerance}
\label{app:tolerance}
\rev{MD observables are estimated from finite trajectories, so a correct simulation reproduces a reference value only up to its statistical uncertainty. For each target quantity, we therefore estimate the standard error $\sigma$ of the reference value, using Flyvbjerg--Petersen blocking~\citep{flyvbjerg1989error} over the production run for time-averaged quantities and the uncertainty of the fit for derived quantities such as Young's modulus. A reported value $x$ is accepted if $|x - x_{\mathrm{ref}}| \le \max(0.05\,|x_{\mathrm{ref}}|,\ 2.77\,\sigma)$. The factor $2.77 = 1.96\sqrt{2}$ gives a two-sided 95\% interval for the difference between two independent estimates that each have standard error $\sigma$, so any value statistically consistent with the reference is accepted. The 5\% floor absorbs differences from rounding and from setup choices that the problem statement leaves open. The same criterion applies to every quantity, to every agent, and to the human evaluators.}

\subsection{\lammps{} vs \gromacs{}}
\label{app:engine_diff}
\lammps{} and \gromacs{} are both molecular dynamics simulation packages, but they target different scientific communities and use cases. \gromacs{} was developed primarily for biomolecular simulations and is highly optimized for proteins, nucleic acids, lipid membranes, and solvated biological systems using fixed-charge force fields like AMBER, CHARMM, GROMOS, and OPLS. Its strength lies in raw performance on standard biomolecular workflows, with aggressive SIMD vectorization and efficient GPU offloading for the kinds of homogeneous systems typical in structural biology and drug discovery. \lammps{}, by contrast, is a general-purpose particle simulator with a much broader scope: it supports atomistic, coarse-grained, mesoscale, and granular models, and ships with a wide range of interatomic potentials including reactive (ReaxFF), bond-order (Tersoff, AIREBO), embedded-atom (EAM), and machine-learning potentials. This makes it the standard tool for materials science, soft matter, polymers, solid-state systems, and any problem where the physics extends beyond classical biomolecular force fields. In short, \gromacs{} is the default choice when simulating biology with established force fields, while \lammps{} is preferred when flexibility across potentials, system types, or custom physics matters more than peak throughput on a narrow class of systems.

\subsection{Example Task Prompt}
\label{app:task_prompt}
\begin{tcolorbox}[
  title=Prompt Used in MDGym to Perform  an Autonomous Simulation,
  colback=gray!5,
  colframe=black,
  breakable,            
  enhanced,             
]
\begin{lstlisting}[
  breaklines=true,
  breakatwhitespace=true,
  basicstyle=\small\ttfamily,
  columns=flexible,
]
Description: Compute the average potential energy per particle for a system of polyethylene (hydrocarbon polymer) using the AIREBO potential. Model the hydrocarbon polymer using atomic units in the metal unit system and an atomic atom style. Construct the system with 60 atoms (20 C and 40 H atoms). Replicate the polyethylene structure 17 times along the x direction, 16 times along y, and 2 times along z to build a sufficiently large bulk system. Construct short range neighbour lists using a binning method with a neighbour cutoff of 0.5 Angstrom, updated every timestep with a delay of five steps. Model interatomic interactions using the AIREBO potential with a Lennard Jones cutoff of 3.0 Angstrom and both Lennard Jones and torsional interactions enabled, using appropriate carbon-hydrogen parameterization. Assign initial velocities corresponding to a temperature of 300 K. Evolve the system under constant-energy (NVE) dynamics with a timestep of 0.0005 ps, and print thermodynamic quantities every 10 steps during a short production run of 1000 timesteps.

Required Metrics: average_potential_energy_per_particle

============================================================
IMPORTANT INSTRUCTIONS:
============================================================
0. You have a total time limit of 3450 seconds to solve this problem. Plan and execute efficiently within this budget.
1. Generate all necessary code, scripts, and input files for the simulation in LAMMPS and write them to the working directory.
2. Format code blocks with the filename for automatic extraction:
   ```language:filename
   code here...
   ```
3. Example formats for LAMMPS:
   - ```lammps:simulation.in
   - ```python:run_simulation.py
   - ```bash:run_simulation.sh
   Example formats for GROMACS:
   - ```gromacs:md.mdp
   - ```bash:run_gromacs.sh
4. The potential name is specified in the problem description. You don't need to download the potential file from the internet; you can access the potential file from the path: /MDGym/data/potentials/36_airebo/.
5. You can access the structure file from the path: /MDGym/data/structures/36_airebo/. If the structure file is not present in that directory, you can make your own structure file based on the information provided in the problem description.
6. You don't need to check the compatibility of the environment or install the MD engine; always assume that the environment is pre-configured with the necessary software and dependencies to run the simulations. Start solving the problem right away as we have a time limit to solve the problem.
7. Run the code and perform any required postprocessing to obtain the final answer.
8. Report the quantities asked in the problem description in JSON format, where the key is the name of the quantity and the value is its numerical value without units.
9. Pay careful attention to units. All required units are specified in the problem description, and the final answer must use those exact units.
10. Your final response must be ONLY a raw JSON object---no markdown, no code fences, no backticks, no explanation, and no preamble. The response must start with { and end with }.
11. Write the final JSON object to final_answer.json for automatic extraction.
12. These are the only instructions. There is no AGENTS.md or such files.
\end{lstlisting}
\end{tcolorbox}

{\begin{tcolorbox}[
  title=Example Level 1 Problem Description,
  colback=gray!5,
  colframe=black,
  breakable,            
  enhanced,             
]
\begin{lstlisting}[
  breaklines=true,
  breakatwhitespace=true,
  basicstyle=\small\ttfamily,
  columns=flexible,
]
Compute the average total energy per nanoparticle for a two-dimensional system of 100 nanoparticles with a number density of 0.104. The simulation is modeled using Lennard-Jones (LJ) units and the body atom style, where each nanoparticle is represented as a rigid collection of constituent points forming a polygonal shape. All particles should be assigned initial velocities from a random distribution corresponding to a reduced temperature of 1.44. Inter-particle interactions are governed by the body/nparticle pair potential with interaction parameters $\epsilon = 1.0$ and $\sigma = 1.0$, using a cutoff distance of 5.0 reduced LJ units. The system should evolve using the nve/body integrator, with an enforce2d constraint applied to restrict all particle motion strictly to the xy-plane. The simulation must be carried out for a total of 100,000 timesteps.
\end{lstlisting}
\end{tcolorbox}}

\begin{tcolorbox}[
  title=Example Level 2 Problem Description,
  colback=gray!5,
  colframe=black,
  breakable,            
  enhanced,             
]
\begin{lstlisting}[
  breaklines=true,
  breakatwhitespace=true,
  basicstyle=\small\ttfamily,
  columns=flexible,
]
Compute the radius of gyration (in nm) of a beta-hairpin peptide (2GB1). The peptide should be solvated with TIP3P water molecules within a cubic simulation box with a side length of 10 nm. The system is to be modeled using the CHARMM27 force field. Energy minimization should be performed using the steepest-descent algorithm with a step size of 0.01 nm for a maximum of 50,000 steps. Following minimization, an NPT equilibration should be conducted. During this phase, long-range electrostatic interactions must be calculated using the Particle Mesh Ewald (PME) method with a real-space cutoff of 1.0 nm, while short-range van der Waals interactions should be treated using a spherical cutoff of 1.0 nm. The temperature should be maintained at 300 K using the velocity-rescale thermostat with a coupling constant of 0.1 ps, and pressure should be maintained at 1 atm using the Parrinello-Rahman barostat with a coupling constant of 2 ps. Bond lengths within the solute must be constrained using the LINCS algorithm. The equilibration run should be conducted for 100 ps with a time step of 2 fs, followed immediately by a 100 ps production run utilizing the exact same parameters.
\end{lstlisting}
\end{tcolorbox}

\begin{tcolorbox}[
  title=Example Level 3 Problem Description,
  colback=gray!5,
  colframe=black,
  breakable,            
  enhanced,             
]
\begin{lstlisting}[
  breaklines=true,
  breakatwhitespace=true,
  basicstyle=\small\ttfamily,
  columns=flexible,
]
Compute the mean square displacement (MSD) and the diffusion coefficient of a 2D Lennard-Jones (LJ) fluid, where the diffusion coefficient. The LJ fluid must be restricted to two dimensions in a simulation box of 40 * 40 (LJ units). Particles should initially be arranged on an sq2 lattice at a number density of 0.6 and interact via the LJ potential with a cutoff radius of 2.5 sigma. All particles have unit mass, and initial velocities must be assigned at a temperature of 1.0. The system is first equilibrated for 5,000 timesteps using NVE integration, with a Langevin thermostat maintaining the temperature. After equilibration, the thermostat should be removed. The production run is then performed in the NVE ensemble for 100,000 timesteps.
\end{lstlisting}
\end{tcolorbox}

\subsection{Dataset details}
\label{app:dataset_details}
The input files used for the {\gromacs} are taken from multiple sources including original research papers, repositories associated tutorials or examples for these packages, or created from scratch. Specifically, 14 files from \gromacs{} repository were taken and modified (with different material, ensemble etc.), 1 file each from \cite{rajput2023ethylene,rajput_2025}, 3 files from \cite{Lemkul_2018}, and 6 files from \cite{gravelle2025tutorials}, and the remaining files are developed by the author. The simulation workflow in {\gromacs} typically follow these steps. First, system is energy minimized using the steepest-descent algorithm. Second, equilibration is performed under NVT and/or NPT ensembles. Finally, production runs are carried out in the NPT ensemble. All observables are computed through post-processing of data obtained from either the equilibration or production runs. 

For the \lammps{} tasks, 73 input files are taken from the official {\lammps} repository~\cite{Thompson_LAMMPS}, 9 files from \cite{gravelle2025tutorials}, and the remaining files are taken from the {\em Mississippi State University} tutorial webpage. All these were subsequently modified with a different material, simulation ensemble or other conditions such as loading or deformation. The simulation workflow in {\lammps} follows a similar protocol unless otherwise specified in the problem description. First, the system is initialized using the provided input parameters. Second, the force field is defined. Third, energy minimization is performed, followed by velocity initialization and equilibration. Finally, production runs are carried out. Unless stated otherwise, all observables are computed by post-processing the production trajectories.

\rev{The knowledge questions are built on the problem statement of an \mdgym{} simulation problem and ask about a decision that the problem requires. The questions were written by the same domain experts who curated the problems with LLM assistance for language correction. We provide a few examples of knowledge questions below.}

\newtcolorbox{questionbox}{
    colback=gray!3,
    colframe=black!35,
    boxrule=0.4pt,
    arc=1mm,
    left=2mm,
    right=2mm,
    top=2mm,
    bottom=2mm,
    breakable
}

\begin{questionbox}
\textbf{Q1.} How should the simulation box be defined in the
$z$-direction for a 2D LAMMPS simulation?

\begin{enumerate}[label=(\Alph*), nosep, leftmargin=1.5em]
    \item Make the $z$-dimension equal to the $x$-dimension.
    \item Make the $z$-dimension equal to the $y$-dimension.
    \item Set the $z$-dimension to a negligible thickness
          (e.g., from $-0.5$ to $0.5$).
    \item Remove the $z$-dimension completely.
\end{enumerate}
\end{questionbox}

\begin{questionbox}
\textbf{Q2.} How is the SLLOD equations of motion algorithm incorporated
in shear flow simulations?

\begin{enumerate}[label=(\Alph*), nosep, leftmargin=1.5em]
    \item Using \texttt{fix nvt/sllod} with \texttt{fix deform}.
    \item Using \texttt{fix nve} with \texttt{fix deposit}.
    \item Using \texttt{fix msst} with \texttt{fix shock}.
    \item Using \texttt{fix press/berendsen} with \texttt{fix rigid}.
\end{enumerate}
\end{questionbox}

\begin{questionbox}
\textbf{Q3.} Which atom style is generally more appropriate for
spherical particles with rotational degrees of freedom?

\begin{enumerate}[label=(\Alph*), nosep, leftmargin=1.5em]
    \item \texttt{atomic}
    \item \texttt{full}
    \item \texttt{sphere}
    \item \texttt{ellipsoid}
\end{enumerate}
\end{questionbox}

\begin{table}[ht]
\centering
\caption{Distribution of \mdgym{} simulation tasks across difficulty levels.}
\label{tab:difficuly_level}
\begin{tabular}{lccc}
\toprule
\textbf{Engine} & \textbf{Easy} & \textbf{Medium} & \textbf{Hard} \\
\midrule
\gromacs{} & 30 & 40 & 30\\
\lammps{} & 34 & 30 & 36\\
\midrule
\textbf{Total} & \textbf{64} & \textbf{70} & \textbf{66} \\
\bottomrule
\end{tabular}
\end{table}

\subsection{Property Categories}
\label{app:property_categories}

\mdgym{} covers a broad range of physical quantities across six property categories, reflecting the diversity of scientific questions that MD simulation is used to answer. Table~\ref{tab:property_categories} summarises the distribution of metrics across categories, and Table~\ref{tab:top_metrics} lists the most frequently occurring individual metrics in the benchmark.

The benchmark is dominated by thermodynamic properties, which appear in the largest number of problems — average temperature, average density, and average potential energy per particle are among the most common individual metrics, reflecting their centrality to standard MD workflows. Structural properties constitute the second-largest category, with coordination number and average radius of gyration being the most frequent. Mechanical and thermodynamic properties are tied as the most diverse categories (24 distinct metrics each). Mechanical properties cover quantities such as Young's modulus, shear modulus, Poisson ratio, and vacancy formation energy that reflect the breadth of materials science applications targeted by \lammps{} tasks. Transport properties (diffusion coefficient, viscosity, thermal conductivity, surface tension) and magnetic properties are represented by a smaller number of problems, providing coverage of more specialised simulation workflows. A small set of numerical validation metrics, such as force relative error and Born matrix relative error, is also included to test whether agents correctly implement numerical verification procedures.

\begin{table}[ht]
\centering
\caption{Distribution of output metrics across property categories in \mdgym{}.}
\label{tab:property_categories}
\begin{tabular}{lcc}
\toprule
\textbf{Property Category} & \textbf{Distinct Metrics} & \textbf{Total Occurrences} \\
\midrule
Thermodynamic   & 24 & 211 \\
\rev{Structural}     & 13 & 94  \\
\rev{Mechanical}      & 24 & 41  \\
Transport       & 6  & 6   \\
Magnetic        & 2  & 3   \\
Other/Numerical & 7  & 11   \\
\midrule
\textbf{Total}  & \textbf{76} & \textbf{366} \\
\bottomrule
\end{tabular}
\end{table}

\begin{table}[ht]
\centering
\caption{Ten Most frequently occurring output metrics in \mdgym{}.}
\label{tab:top_metrics}
\begin{tabular}{lcc}
\toprule
\textbf{Metric} & \textbf{Category} & \textbf{Problems} \\
\midrule
Average temperature               & Thermodynamic & 114 \\
Coordination number               & Structural    & 49 \\
Average density                   & Thermodynamic & 41 \\
Average radius of gyration        & Structural    & 16 \\
Average potential energy per particle & Thermodynamic & 13 \\
Average total energy              & Thermodynamic & 11  \\
Average total energy per particle & Thermodynamic & 9 \\
Young's modulus                   & Mechanical    & 8  \\
Average end-to-end distance       & Structural    & 7  \\
Solvent Accessible Surface Area & Structural & 6 \\
\bottomrule
\end{tabular}
\end{table}

\subsection{MD Simulation Details}
Molecular dynamics (MD) simulations were carried out using two widely used MD packages: GROMACS and LAMMPS. Specifically, GROMACS v2026.0 package and LAMMPS (22 Jul 2025 version) were used to perform the simulations. To enable simulation execution through LLM models, we provided the initial system structures along with the corresponding topological parameters.

\section{Agent Framework Architectures}
\label{app:agent_architectures}

To characterize precisely what kind of system each evaluated framework is, this section documents the publicly available architectural details of Claude Code, Codex, and OpenHands along four criteria relevant to general-purpose agentic capability: system prompt structure, state management, tool descriptions, and error recovery mechanisms. We report only what is documented or inspectable in each framework's public materials, without interpretation.

\subsection{Claude Code}

Claude Code's documented architecture is an agentic loop of three phases --- gather context, take action, verify results --- in which the model determines the requirements of each step from the outcome of the previous one, chaining actions and course-correcting as needed\footnote{\url{https://code.claude.com/docs/en/how-claude-code-works}}. The official documentation states explicitly: \textit{``While it excels at coding, it can help with anything you can do from the command line: writing docs, running builds, searching files, researching topics, and more.''} State is persisted via a project-level \texttt{CLAUDE.md} file and auto-loaded memory. The framework additionally exposes a public glossary defining its \textit{agentic harness} --- ``the tools, context management, and execution environment that turn a language model into a capable coding agent'' --- as distinct from the underlying model.

\subsection{Codex}

Codex's documented architecture centers on a shared \textit{harness} --- the core agent loop and execution logic underlying Codex CLI, Codex Cloud, and the Codex VS Code extension. The loop follows a standard inference-then-tool-call pattern: user input is incorporated into a prompt sent to the Responses API; the model either produces a final assistant message or requests a tool call, which the harness executes and appends to the prompt before re-querying the model; this repeats until an assistant message terminates the turn. The default tool set includes a \texttt{shell} tool for spawning local processes, a built-in \texttt{update\_plan} tool, a \texttt{web\_search} tool provided by the Responses API, and user-configured MCP server tools. State and task-specific instructions are aggregated at prompt-construction time from \texttt{AGENTS.override.md} and \texttt{AGENTS.md} files found from the project root down to the working directory, analogous to Claude Code's \texttt{CLAUDE.md}. Codex additionally supports \textit{skills}, which OpenAI describes as extending Codex ``beyond code generation to tasks that require gathering and synthesizing information, problem-solving, writing, and more''\footnote{\url{https://openai.com/index/introducing-the-codex-app/}}. We note, however, that OpenAI's own product framing of Codex is more coding-centric than Claude Code's: OpenAI's developer guidance recommends general-purpose models for non-coding work and positions ``the Codex surface'' specifically for tasks involving ``file edits, shell commands, reviews, and sandboxes'', and third-party summaries describe Codex as ``an AI coding agent... for software engineering tasks.''

\subsection{OpenHands}

OpenHands' system prompt is assembled from a typed, open-source section registry comprising sixteen sections in its base configuration: ROLE, MEMORY, EFFICIENCY, FILE\_SYSTEM\_GUIDELINES, CODE\_QUALITY, VERSION\_CONTROL, PULL\_REQUESTS, PROBLEM\_SOLVING\_WORKFLOW, SELF\_DOCUMENTATION, SECURITY, SECURITY\_RISK\_ASSESSMENT, EXTERNAL\_SERVICES, ENVIRONMENT\_SETUP, TROUBLESHOOTING, PROCESS\_MANAGEMENT, and an optional model-specific IMPORTANT section, followed by dynamic per-conversation content. The default tool set consists of \texttt{TerminalTool}, \texttt{FileEditorTool}, \texttt{TaskTrackerTool}, and \texttt{BrowserToolSet} (disabled in CLI mode), with \texttt{FinishTool} and \texttt{ThinkTool} attached automatically as additional built-ins\footnote{\url{https://github.com/OpenHands/software-agent-sdk}}.

Of the sixteen base sections, three --- CODE\_QUALITY, VERSION\_CONTROL, and PULL\_REQUESTS --- carry explicit software-engineering framing. We report this for transparency: these sections are present in the default configuration and were not disabled during evaluation. We also note that OpenHands' own project documentation self-describes with an explicit coding emphasis: its repository README states \textit{``Emphasis on coding: Purpose-built for software development tasks with specialized tools and workflows''}. Of the three frameworks evaluated, this is the most explicitly coding-scoped self-description; Claude Code's documentation makes the opposite claim (general-purpose, coding as one use case among several), and Codex's positioning falls between the two, offering general-purpose extension via skills while being marketed primarily as a coding surface.

\subsection{Cross-Framework Comparison}

Table~\ref{tab:agent_architecture_criteria} summarizes how each of the four architectural criteria is instantiated across the evaluated harnesses.

\begin{table}[ht]
\centering
\small
\caption{Architectural components of evaluated agent frameworks, mapped to four criteria relevant to general-purpose agentic capability.}
\label{tab:agent_architecture_criteria}
\begin{tabular}{lp{9.5cm}}
\toprule
\textbf{Criterion} & \textbf{Instantiation in evaluated harnesses} \\
\midrule
System prompts & Typed section registry (OpenHands); model-specific bundled instructions read from \texttt{config.toml} or built-in defaults (Codex); documented three-phase agentic loop (Claude Code) \\
State management & MEMORY section and \texttt{TaskTrackerTool} (OpenHands); \texttt{AGENTS.md}/\texttt{AGENTS.override.md} aggregation (Codex); \texttt{CLAUDE.md} and auto-loaded memory (Claude Code) \\
Tool descriptions & JSON-schema function-call tools (\texttt{shell}, \texttt{update\_plan}, \texttt{web\_search}, MCP) (Codex); typed tool classes (\texttt{TerminalTool}, \texttt{FileEditorTool}, etc.) (OpenHands); documented built-in tool set (Claude Code) \\
Error recovery loops & TROUBLESHOOTING and PROBLEM\_SOLVING\_WORKFLOW sections (OpenHands); the verify phase of the agentic loop (Claude Code); iterative tool-call/re-inference cycle with automatic context compaction (Codex) \\
\bottomrule
\end{tabular}
\end{table}

\subsection{Discussion}
We report this distribution and self-description transparently rather than omitting it, and note that it applies unevenly across the three frameworks: Claude Code's documentation makes the strongest explicit claim to general-purpose, non-coding capability, while OpenHands' own materials make the strongest explicit claim to coding-specific purpose-building. This observation does not, on its own, establish that \mdgym{} performance is measuring software-engineering-specific priors rather than general agentic capability. The architectural machinery common to all three evaluated frameworks --- a terminal, a file editor, a task-tracking mechanism, a documented troubleshooting or verification protocol, and state persisted via a project-level instructions file (\texttt{CLAUDE.md} / \texttt{AGENTS.md} / \texttt{MEMORY.md}) --- is precisely the machinery required to operate any MD engine, since MD workflows are executed through command-line binaries acting on text-based input decks.

\section{Error Analysis Methodology}
\label{app:error_modes}

\subsection{Trajectory-Level Error Taxonomy}

The trajectory-level error taxonomy was developed through iterative manual inspection of failed agent trajectories. We began by reviewing a random sample of failed runs across all four models, identifying recurring patterns in how agents fail. This process yielded five mutually non-exclusive error categories — a trajectory may exhibit multiple error types simultaneously. For each category, we define both the operational definition and the rule used for automated classification. \rev{The error detection rules were developed on the basis of 20 failed trajectories and then frozen for the rest of the experiments. A domain expert (one of the authors) independently labeled a separate sample of 250 failed trajectories, which agreed with all 250.}

\rev{\paragraph{Fabricated Answer.} The agent reports numerical values in \texttt{final\_answer.json} that were not obtained from a completed simulation. Either no simulation was run, or every simulation the agent ran terminated with an engine error before producing the required output. The reported values are therefore guessed, recalled, or estimated rather than computed, even when they look plausible. This category differs from \emph{Incorrect answer from error-free run}, in which a simulation completed without errors but the reported value is wrong. \\
\textit{Detection rule:} The agent wrote \texttt{final\_answer.json}, but either no simulation output files exist in its working directory, or the log of the agent's last simulation run ends with an engine error, so that no simulation completed.}

\paragraph{Syntax Error.} The agent produces malformed input scripts that prevent the simulation from executing. \\
\textit{Detection rule:} A failed agent action where the action output contains a syntax-indicating keyword such as \texttt{syntax error}, \texttt{unknown command}, \texttt{invalid}, \texttt{illegal}, \texttt{unrecognized}, or \texttt{fatal error} alongside an engine-specific marker such as \texttt{lmp}, \texttt{.in}, \texttt{gmx}, \texttt{.mdp}, or \texttt{.top}.

\paragraph{Environment Misunderstanding.} The agent fails to correctly interpret the task setup, available tools, or the pre-configured simulation environment, leading to actions that are irrelevant or counterproductive to the task. \\
\textit{Detection rule:} A failed action whose output contains \texttt{command not found}.

\paragraph{Premature Termination.} The agent abandons the task upon encountering domain-specific error messages, failing to attempt iterative debugging or recovery. \\
\textit{Detection rule:} The agent produced no answer and the run did not exhaust the time budget. Timed-out runs are excluded, as failing to produce an answer under a binding time constraint is expected behaviour rather than premature termination.

\rev{\paragraph{Incorrect Post-Processing.} The agent executes the simulation but applies an incorrect analysis procedure to extract the required output metric, producing a wrong final value despite a physically valid simulation. \\
\textit{Detection rule:} The agent simulation log produces correct values when the golden post-processing script is used.}

\rev{\paragraph{Incorrect Simulation.} The agent executes an error-free simulation and produce a wrong final value despite using correct post processing. \\
\textit{Detection rule:} The agent simulation log produces incorrect values when the golden post-processing script is used.}

\rev{\paragraph{Failure Despite Error-Free Run.} The agent produces an answer but still fail, either due to incorrect post-processing or incorrect simulation. \\
\textit{Detection rule:} The agent produced an answer, and the simulation ran without errors, ruling out fabrication, but the trajectory still failed.}

\subsection{Simulation-Level Error Classification}

The simulation-level error analysis characterizes the specific errors produced by the MD engines during failed simulation runs, providing a finer-grained view of where and how simulations break down. Error categories for each engine were predefined by domain experts based on the known failure modes of \lammps{} and \gromacs{}:

\paragraph{\lammps{} categories} Command Syntax, Variable/Expression, Pair/Force Field, Data File and Read, Fix/Compute Incompatibility, Thermo and Output, Unit and Dimension, Region and Geometry, Package/Partition, Lost Atoms, Energy Divergence, Neighbor List, Timestep, Script Logic Order, Boundary Condition, Thermostat/Barostat, MPI/Parallel Decomposition, and Constraint Failure.

\paragraph{\gromacs{} categories} MDP Parameter, Integrator/Thermostat Incompatibility, Topology, Index Group, Coord/Topology Mismatch, Position Restraint, LINCS/Constraint, Energy Blow-up, Periodic Boundary/Box Error, PME/Electrostatics, MPI/GPU Parallelisation, Force Field Param Missing, Velocity/Temperature Initialisation, Pressure Coupling/Barostat, and Output/File Write.

For each failed simulation run, engine error logs were extracted from the trajectory and provided to Claude Sonnet 5 along with the predefined category list and their definitions. To improve robustness, three independent classifications were generated per error and the majority label was taken as the final category. \rev{The three LLM-as-a-judge votes were unanimous for over 95\% of engine errors, and the expert agrees with every unanimous case checked. All the non-unanimous cases involved the None and Unclassified categories. Further, we used Claude Opus 5.5 to independently re-classify 250 engine errors using the same definitions. The two models disagreed on about 4\% of errors, all in the MPI / Parallel category, which is 1.2\% of engine errors.} The system prompts used for \lammps{} and \gromacs{} classification are provided below.

\subsection{LLM Classifier Prompts}
\label{app:judge_prompt}

\subsubsection{\lammps{} Error Classifier Prompt}

\begin{tcolorbox}[
  title=LAMMPS Error Classifier System Prompt,
  colback=gray!5,
  colframe=black,
  breakable,            
  enhanced,             
]
\begin{lstlisting}[
    breaklines=true, 
    breakatwhitespace=true, 
    basicstyle=\small\ttfamily, 
    columns=fullflexible,
    breakindent=0pt
]
You are a LAMMPS simulation error classifier. You will be given a list of numbered error messages extracted from a LAMMPS trajectory. For each error, classify it into exactly one category from the taxonomy below.

=== SYNTAX / STATIC ERRORS (S1-S9) ===
These are detected during input script parsing before any timestep runs.

--- S1: Command Syntax Error ---
Definition: A misspelled LAMMPS command name, incorrect number of arguments, arguments in the wrong order, or use of a keyword that does not exist in the installed build. Detected during input script parsing before any timestep runs.
Trigger: Parser fails to recognise the command token or cannot match the argument list to any known signature for that command.
Signals: 'Unknown command', 'Illegal command', or 'Expected N arguments'; error line references the exact input script line; error occurs before timestep 0.
Classification rule: Classify as S1 if ALL criteria hold: (1) LAMMPS reports 'Unknown command', 'Illegal command', 'Expected N arguments', or 'Unrecognized keyword'; (2) the error is raised during input script parsing before the run starts; (3) the message cites a specific script line.
Examples: pair_style lj/cut 2.5 extra_arg  # too many args; atom_sytle atomic  # typo in command name

--- S2: Variable & Expression Error ---
Definition: Use of an undefined variable, an invalid mathematical expression inside a variable definition, or the wrong variable style for a context.
Trigger: LAMMPS tries to evaluate a variable or expression and cannot resolve the reference or parse the math.
Signals: 'Variable ... is not defined'; 'Invalid math expression'; 'Variable ... has wrong style'.
Classification rule: Classify as S2 if ALL criteria hold: (1) the message contains 'variable', 'undefined variable', 'invalid expression', or 'wrong style'; (2) the failure occurs during variable evaluation or preprocessing; (3) no atom positions or forces are involved.

--- S3: Pair / Force Field Error ---
Definition: Mismatch between pair_style and pair_coeff declarations: missing coefficients for one or more atom-type pairs, wrong number of parameters, inconsistent mixing rules, or use of a pair style not compiled into the LAMMPS binary.
Signals: 'No pair style is defined'; 'Pair coeffs are not set'; 'Incorrect args for pair_coeff'.
Classification rule: Classify as S3 if ALL criteria hold: (1) the message references 'pair_style', 'pair_coeff', 'pair coeffs', or 'mixing'; (2) the error appears before or at the start of the run; (3) it does not reference lost atoms or energy divergence.

--- S4: Data File & Read Error ---
Definition: Malformed LAMMPS data file: atom count in the header does not match the Atoms section, a required section is absent, atom type indices are out of range, or the file format does not match the declared atom_style.
Signals: 'Did not assign all atoms correctly'; 'Unexpected end of data file'; 'Too many/few lines in ... section'; 'Invalid atom type'.
Classification rule: Classify as S4 if ALL criteria hold: (1) the error is raised inside or immediately after read_data; (2) it references atom counts, sections, or file structure; (3) it does not reference pair coefficients or force constants.

--- S5: Fix / Compute Incompatibility ---
Definition: A fix or compute applied to a group that does not exist, references a compute/fix ID not yet defined, or has invalid arguments.
Signals: 'Group ID does not exist'; 'Fix requires ... atom style'; 'Compute ID ... does not exist'.
Classification rule: Classify as S5 if ALL criteria hold: (1) the message names a fix or compute ID; (2) it references a missing group, missing dependency, or incompatible atom style; (3) the error is raised before the first timestep.

--- S6: Thermo & Output Error ---
Definition: Error in thermo_style, dump, or restart configuration: referencing a compute or fix keyword not defined, or mismatching dump format arguments.
Signals: 'Unknown keyword in thermo_style'; 'Invalid keyword in dump ... command'.
Classification rule: Classify as S6 if ALL criteria hold: (1) the message involves thermo_style, dump, or restart; (2) it references an unrecognised or undefined keyword/ID; (3) it is not a runtime energy or force error.

--- S7: Unit & Dimension Error ---
Definition: Commands that depend on a unit system issued before the units command is set, or 2D-specific commands used without declaring dimension 2.
Signals: 'Cannot use ... before units are set'; 'Must use dimension 2 before ...'.
Classification rule: Classify as S7 if ALL criteria hold: (1) the message references 'units', 'dimension', or a unit-dependent quantity; (2) the error appears before the run starts; (3) it does not concern atom types or pair coefficients.

--- S8: Region & Geometry Error ---
Definition: Wrong keyword for a region shape type, incorrect number of geometric parameters, invalid simulation box dimensions, or create_atoms referencing an undefined region.
Signals: 'Illegal region command'; 'Invalid region ID'; 'Box bounds are invalid'.
Classification rule: Classify as S8 if ALL criteria hold: (1) the message references 'region', 'box', or 'create_atoms'; (2) the error involves geometric parameters or undefined region IDs; (3) it occurs before the run.

--- S9: Package / Partition Error ---
Definition: Requesting a GPU, KOKKOS, INTEL, or OMP accelerated style not compiled into the LAMMPS binary, or supplying invalid arguments to the package command.
Signals: 'Package gpu is not available'; 'Unrecognized package name'; style suffix /gpu or /kk present but build lacks that suffix.
Classification rule: Classify as S9 if ALL criteria hold: (1) the message references 'package', a style suffix (/gpu, /kk, /intel, /omp), or 'partition'; (2) it states the feature is unavailable or the command is illegal; (3) it occurs at startup before setup.

=== RUNTIME / LOGICAL ERRORS (R1-R9) ===
These occur during a run at timestep > 0.

--- R1: Lost Atoms Error ---
Signals: 'Lost atoms: ...' during the run; 'Atoms lost in xxx timesteps'.
Classification rule: Classify as R1 if ALL criteria hold: (1) the message contains 'lost atoms' or 'atoms lost'; (2) the error is raised during a run (timestep > 0); (3) a count of lost atoms or a timestep number is reported.

--- R2: Energy Divergence Error ---
Signals: Energy printed as NaN or Inf in thermo output; 'Numerical instability' warning.
Classification rule: Classify as R2 if ALL criteria hold: (1) thermo output shows NaN, Inf, or astronomically large energy values; (2) the error occurs during a run (timestep > 0); (3) no 'lost atoms' message appears before the NaN.

--- R3: Neighbor List Error ---
Signals: 'Dangerous builds' counted in output; 'Neighbor list overflow'.
Classification rule: Classify as R3 if ALL criteria hold: (1) the message references 'neighbor', 'neigh', 'skin', or 'dangerous builds'; (2) it appears during setup or at a rebuild step; (3) it does not report lost atoms.

--- R4: Timestep Error ---
Signals: Energy divergence or lost atoms within the first ~50 steps; extremely high temperature in the first few steps.
Classification rule: Classify as R4 if ALL criteria hold: (1) the instability begins within the first ~50 steps; (2) decreasing the timestep resolves the issue; (3) there is no obvious initial geometry overlap.

--- R5: Ordering / Script Logic Error ---
Signals: 'Pair style not yet defined'; 'Must define atom_style before read_data'.
Classification rule: Classify as R5 if ALL criteria hold: (1) the message says a required precondition has not been set; (2) re-ordering the commands would fix the error; (3) no atom positions or forces are involved.

--- R6: Boundary Condition Error ---
Signals: 'Pair style requires periodic boundary in all dimensions'; lost atoms attributed to non-periodic boundary.
Classification rule: Classify as R6 if ALL criteria hold: (1) the message references 'boundary', 'periodic', 'shrink-wrap', or 'fixed wall'; (2) the pair style or ensemble conflicts with the boundary setting; (3) the error does not arise from atom geometry or overlaps.

--- R7: Thermostat / Barostat Error ---
Signals: 'Fix npt requires ...'; pressure or temperature oscillating wildly.
Classification rule: Classify as R7 if ALL criteria hold: (1) the message references a thermostat or barostat fix; (2) the error involves incompatible settings or T/P instability; (3) it is not a lost-atom or energy-divergence error.

--- R8: MPI / Parallel Decomposition Error ---
Signals: 'Too many processors for simulation'; 'Processor grid does not match domain'.
Classification rule: Classify as R8 if ALL criteria hold: (1) the message references 'processors', 'MPI', 'domain', or 'ranks'; (2) the error is triggered by the number of MPI processes or processor grid; (3) it does not involve pair coefficients or atom types.

--- R9: Constraint Failure ---
Signals: 'SHAKE atoms missing'; 'RATTLE failed to converge'; 'Linear constraint solver did not converge'.
Classification rule: Classify as R9 if ALL criteria hold: (1) the message contains 'SHAKE', 'RATTLE', 'LINCS', or 'constraint'; (2) it references convergence failure or missing atoms; (3) it is a runtime error (timestep > 0).

=== INSTRUCTIONS ===
For each numbered error below, return exactly one category ID from: S1 S2 S3 S4 S5 S6 S7 S8 S9 R1 R2 R3 R4 R5 R6 R7 R8 R9 or "Unclassified".

Respond with a valid JSON object only - no markdown, no explanation:
{
  "classifications": [
    {"index": 1, "category_id": "S3", "category_name": "Pair / Force Field Error", "category_type": "Syntax"},
    ...
  ]
}
\end{lstlisting}
\end{tcolorbox}

\subsubsection{\gromacs{} Error Classifier Prompt}

\begin{tcolorbox}[
  title=GROMACS Error Classifier System Prompt,
  colback=gray!5,
  colframe=black,
  breakable,            
  enhanced,             
]
\begin{lstlisting}[
    breaklines=true, 
    breakatwhitespace=true, 
    basicstyle=\small\ttfamily, 
    columns=fullflexible,
    breakindent=0pt
]
You are a GROMACS simulation error classifier. You will be given a list of numbered error messages extracted from a GROMACS trajectory. For each error, classify it into exactly one category from the taxonomy below.

=== PREPROCESSING / STATIC ERRORS (P1-P6) ===
These are detected by grompp before the simulation starts.

--- P1: MDP Parameter Error ---
Definition: An unrecognised, misspelled, or deprecated parameter key in the .mdp run-input file, or a value out of the allowed range.
Signals: Warning: Unknown left-hand 'xyz' in parameter file; Fatal error: Invalid value for mdp option.
Classification rule: Classify as P1 if ALL criteria hold: (1) the message references an .mdp file option by name; (2) it says the option is 'unknown', 'invalid', 'deprecated', or out of range; (3) the error is raised by grompp before the simulation starts.

--- P2: Integrator / Thermostat Incompatibility ---
Definition: An incompatible combination of integrator and thermostat or barostat in the .mdp file.
Signals: Fatal error: Integrator ... is not compatible with tcoupl ...; Fatal error: Cannot use ... with integrator ...
Classification rule: Classify as P2 if ALL criteria hold: (1) the message names both an integrator and a thermostat or barostat; (2) it says they are incompatible; (3) the error is raised by grompp.

--- P3: Topology Error ---
Definition: An error in the .top or .itp topology file: broken #include chain, mismatched moleculetype declarations, wrong atom count, undefined atom types, or missing force field files.
Signals: Fatal error: No such file or directory (for included .itp); Fatal error: number of atoms in ... does not match.
Classification rule: Classify as P3 if ALL criteria hold: (1) the message references a .top or .itp file; (2) it reports a missing include, undefined atom type, or atom count mismatch; (3) it is raised by grompp.

--- P4: Index Group Error ---
Definition: A group name referenced in the .mdp file does not exist in the .ndx index file, or the index file is malformed or missing.
Signals: Fatal error: Group ... not found in index file; Error reading index file.
Classification rule: Classify as P4 if ALL criteria hold: (1) the message references a group name or index file; (2) it says the group is 'not found' or the index file is invalid/missing; (3) it is raised by grompp.

--- P5: Coordinate / Topology Mismatch ---
Definition: The number or order of atoms in the coordinate file does not match what the topology expects.
Signals: Fatal error: number of coordinates in ... does not match topology; Fatal error: Atom ... not found in residue.
Classification rule: Classify as P5 if ALL criteria hold: (1) the message compares a coordinate file to the topology; (2) it reports a count or naming mismatch; (3) it is raised by grompp.

--- P6: Position Restraint Error ---
Definition: The position restraint file does not match the current topology: wrong number of restrained atoms, mismatched atom indices, or the posre.itp #include is missing.
Signals: Fatal error: Position restraint file ... has ... entries, expected ...; Fatal error: Atom index ... in position restraints is out of range.
Classification rule: Classify as P6 if ALL criteria hold: (1) the message references a position restraint file or the posres directive; (2) it reports a count or index mismatch; (3) it is raised by grompp.

=== RUNTIME / LOGICAL ERRORS (R1-R9) ===
These occur during mdrun at step >= 0.

--- R1: LINCS / Constraint Failure ---
Signals: LINCS WARNING: ... constraints not satisfied; Fatal error: Too many LINCS warnings.
Classification rule: Classify as R1 if ALL criteria hold: (1) the message contains 'LINCS', 'SETTLE', or 'constraint'; (2) it references non-convergence or deviation during the run; (3) it occurs at a specific timestep (step > 0).

--- R2: Energy Blow-up / Numerical Instability ---
Signals: Fatal error: Nan or Inf in coordinates or velocities; Step ...: NAN detected in ...
Classification rule: Classify as R2 if ALL criteria hold: (1) the message contains 'NaN', 'Inf', 'blow-up', or 'not converging'; (2) it is raised during an mdrun step (step > 0); (3) no LINCS warning appears before it.

--- R3: Periodic Boundary / Box Error ---
Signals: Fatal error: The box is too small for the largest cut-off; Fatal error: Atom ... is outside the box.
Classification rule: Classify as R3 if ALL criteria hold: (1) the message references the simulation box, cutoff, or PBC; (2) it reports atoms outside the box or box too small; (3) it is a runtime error (step >= 0).

--- R4: PME / Electrostatics Error ---
Signals: Fatal error: The PME grid ... is not compatible with ...; Fatal error: Cannot use PME with non-periodic boundary.
Classification rule: Classify as R4 if ALL criteria hold: (1) the message references 'PME', 'Ewald', 'grid', or 'electrostatics'; (2) it reports an incompatibility or initialisation failure; (3) it does not reference LINCS or NaN.

--- R5: MPI / GPU Parallelisation Error ---
Signals: Fatal error: Too few ranks ... for domain decomposition; Fatal error: Cannot use GPU with ...
Classification rule: Classify as R5 if ALL criteria hold: (1) the message references 'ranks', 'domain decomposition', 'GPU', or 'MPI'; (2) it reports an incompatibility or setup failure; (3) it occurs at initialisation (before step 1).

--- R6: Force Field Parameter Missing ---
Signals: Fatal error: No parameters for ... in ...; Fatal error: No default ... parameters.
Classification rule: Classify as R6 if ALL criteria hold: (1) the message says 'no parameters', 'no default', or 'not found' for a bonded or non-bonded interaction; (2) it is raised during mdrun (step >= 0); (3) it is not a topology #include or atom-count error.

--- R7: Velocity / Temperature Initialisation Error ---
Signals: Fatal error: gen-temp must be > 0; WARNING: Velocities were not generated.
Classification rule: Classify as R7 if ALL criteria hold: (1) the message references 'gen-vel', 'gen-temp', 'velocities', or temperature initialisation; (2) it reports an invalid temperature or velocity incompatibility; (3) it occurs at startup (step 0 or grompp).

--- R8: Pressure Coupling / Barostat Error ---
Signals: WARNING: Step ..., time ...: the box is growing/shrinking rapidly; Fatal error: Box has collapsed.
Classification rule: Classify as R8 if ALL criteria hold: (1) the message references 'pressure coupling', 'barostat', 'box vectors', or 'compressibility'; (2) it reports rapid box change, collapse, or incompatibility; (3) it is a runtime error (step > 0).

--- R9: Output / File Write Error ---
Signals: Fatal error: Cannot write to file ...; Fatal error: Incompatible checkpoint file.
Classification rule: Classify as R9 if ALL criteria hold: (1) the message references a file path or file type (.trr, .xtc, .edr, .cpt); (2) it says the file cannot be written, is corrupt, or is incompatible; (3) it is raised during mdrun output.

=== INSTRUCTIONS ===
For each numbered error below, return exactly one category ID from: P1 P2 P3 P4 P5 P6 R1 R2 R3 R4 R5 R6 R7 R8 R9 or "Unclassified".

Respond with a valid JSON object only - no markdown, no explanation:
{
  "classifications": [
    {"index": 1, "category_id": "P3", "category_name": "Topology Error", "category_type": "Preprocessing"},
    ...
  ]
}
\end{lstlisting}
\end{tcolorbox}

\begin{tcolorbox}[
  title=Simulation Stage Completion Classifier System Prompt,
  colback=gray!5,
  colframe=black,
  breakable,            
  enhanced,             
]
\begin{lstlisting}[
    breaklines=true, 
    breakatwhitespace=true, 
    basicstyle=\small\ttfamily, 
    columns=fullflexible,
    breakindent=0pt
]
You are an expert computational materials scientist and an automated diagnostic engine for molecular dynamics (MD) simulations. 

Your task is to analyze the provided simulation log file (which may be from LAMMPS, GROMACS, or similar engines) and determine the exact stage at which the simulation ended, whether it completed successfully, or if it failed, what type of error caused the failure.

### The 4 Stages of MD Simulation (Strict Definitions)
Use the following explicit criteria to classify the progress of the simulation. Do not guess; rely on these strict physical and textual signatures:

1. INITIALIZATION (System Setup)
- Definition: The engine is translating inputs into a mathematical state prior to any particle displacement. Time is strictly at t=0. This includes parsing topology, applying forcefield parameters (charges, LJ parameters), building initial neighbor lists, and allocating memory.
- Success Signatures: The log explicitly prints the simulation box dimensions, total number of atoms/interactions, and successfully evaluates step 0 (or pre-run) energies without throwing "NaN" (Not a Number) or infinite force errors.
- Exclusion: If the engine crashes before building the neighbor list or printing step 0 thermals, Initialization has FAILED.

2. ENERGY MINIMIZATION (Static Relaxation)
- Definition: A purely mathematical optimization to find a local potential energy minimum and resolve steric clashes. There is no concept of time, temperature, or kinetic energy in this stage. Algorithms used are typically Steepest Descent (SD) or Conjugate Gradient (CG).
- Success Signatures: The log explicitly reports "Minimization converged", the maximum force (F_max) drops below the target tolerance (e.g., ftol), or the maximum requested iterations are safely reached. 
- Exclusion: If atoms move "too far" or forces become infinite here, it is a Minimization failure.

3. THERMODYNAMIC EQUILIBRATION (NVT / NPT)
- Definition: Time-integration begins (Newton's equations are solved), but the primary goal is coupling the system to a heat bath (thermostat) or pressure bath (barostat) to reach a target macrostate. Trajectory data here is considered "burn-in" and not meant for final analysis.
- Success Signatures: Completion of the specified NVT/NPT `run` segments. The log must show rolling averages of Temperature, Pressure, and Density oscillating steadily around their target values (e.g., T ~ 300K) with no exponential drift or sudden spikes.
- Exclusion: If a `run` command finishes, but the final temperature is thousands of degrees away from the target, Equilibration has FAILED due to physical blow-up, even if the engine didn't explicitly crash.

4. PRODUCTION RUN (Data Acquisition)
- Definition: The final, stable integration phase where the thermodynamic ensemble is maintained, and coordinates/velocities are actively dumped to trajectory files for scientific analysis. 
- Success Signatures: The final `run` command completes entirely. The log prints the ultimate timing summary (e.g., "Total wall time", "Performance: ns/day", "Loop time of X on Y procs"), signaling a clean exit of the simulation loop.
- Exclusion: Any crash, lost atom, or constraint failure occurring after equilibration completes is a Production failure.

### Task Instructions
1. Scan the log file sequentially from top to bottom.
2. Identify the highest stage that completed successfully. If a stage started but crashed before finishing, the *previous* completed stage is the last successful one. If it crashes during Initialization, the last successful stage is "None".
3. If the simulation crashed, classify the failure using one of these categories: [Syntax/Parsing, Topology/Parameterization, Step-Zero Instability, Integration/Dynamics Instability, Hardware/Parallelization].
4. Extract the specific error message or the line right before the failure.

### Output Format
You must output your analysis as a valid JSON object using the exact schema below. Do not include any markdown formatting or conversational text outside of the JSON block.

{
  "last_successful_stage": "None" | "Initialization" | "Minimization" | "Equilibration" | "Production",
  "error_category": "String (from the categories listed above, or null if Success)",
  "error_evidence": "String (the exact error message or last printed line indicating the crash, or null if Success)",
  "reasoning": "A brief, 1-2 sentence explanation justifying the classification."
}
\end{lstlisting}
\end{tcolorbox}

\subsection{Agent Behavior Modes}

During manual inspection of agent trajectories, we observed recurring behavioral patterns across agents. We document these here as qualitative observations to complement the automated error analysis.

\textbf{Plan Then Execute.} The agent produces substantial reasoning about the problem — identifying required parameters, decomposing sub-tasks, and anticipating potential pitfalls before writing any simulation files.

\textbf{Exploratory Probing.} The agent begins by exploring the environment, for example, listing files, checking directory contents, reading existing structure or potential files, before writing any simulation code.

\textbf{Code Generation.} The agent generates a simulation input script, either writes the first file immediately after reading the problem or produces all required files in a single batch before executing any of them.

\textbf{Iterative Refinement.} The agent writes a simulation file, executes it, encounters an error, edits the file, and retries — repeating this write-run-debug cycle. The same core approach is preserved and incrementally corrected rather than abandoned.

\textbf{Trial and Error.} The agent makes multiple substantially different attempts to solve the problem, abandoning one approach for another when it fails. Distinguished from Iterative Refinement by the scope of change between attempts: the agent rewrites the simulation strategy rather than patching a specific error.

\textbf{Verification.} The agent explicitly checks whether its simulation output is correct before submitting a final answer, including parsing output logs to extract specific metric values, cross-referencing computed results against expected physical ranges, and performing unit conversions on extracted values.

\section{Additional error analysis}
\begin{figure}[ht]
    \centering
    \vspace{-0.1in}
    \includegraphics[width=\linewidth]{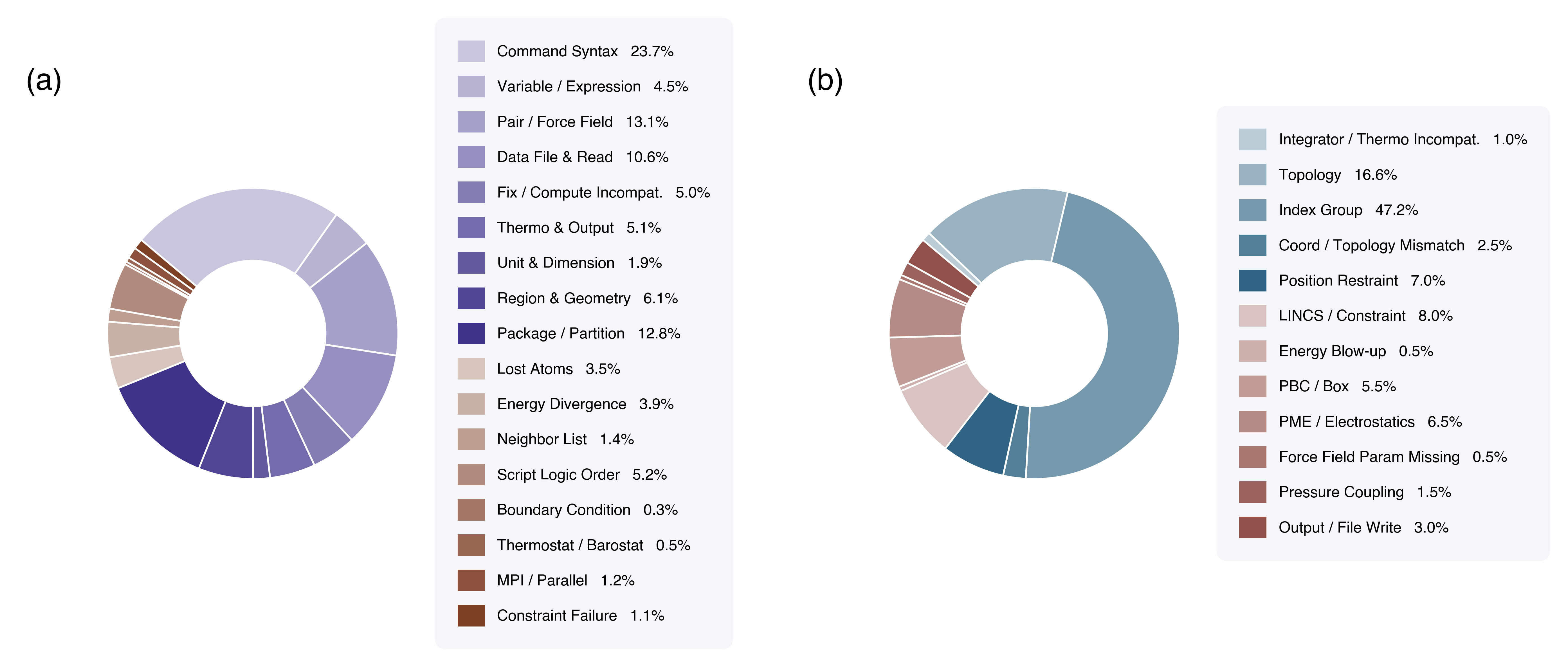}
    \vspace{-0.1in}
    \caption{Distribution of engine errors raised within failed trajectories, pooled over all configurations, for (a) \lammps{} and (b) \gromacs{}. Categories follow App.~\ref{app:error_modes}.}
    \vspace{-0.1in}
    \label{fig:engine_specific_errors}
\end{figure}

\begin{table}[ht]
\centering
\small
\caption{\rev{Prevalence of failure modes, as the percentage
of failed trajectories exhibiting each mode. \% indicates the number of trajectories having a failure mode; one trajectory can have multiple failure modes.}}
\label{tab:failure_categories}
\setlength{\tabcolsep}{4pt}
\begin{tabular}{l>{\centering\arraybackslash}p{1.6cm}>{\centering\arraybackslash}p{1.5cm}>{\centering\arraybackslash}p{1.9cm}>{\centering\arraybackslash}p{1.6cm}>{\centering\arraybackslash}p{1.6cm}>{\centering\arraybackslash}p{1.2cm}}
\toprule
\textbf{Model} & \textbf{Env.\ Misunder\-standing} & \textbf{Fabricated Answer} & \textbf{Incorrect Post-processing} & \textbf{Incorrect Simulation} & \textbf{Premature Termination} & \textbf{Syntax Error} \\
\midrule
Claude Opus 4.8 & 0\% & 1\% & 3\% & 18\% & 15\% & 30\% \\
GPT 5.5         & 0\% & 6\% & 7\% & 31\% & 4\%  & 43\% \\
DeepSeek V4 Pro & 5\% & 3\% & 2\% & 4\%  & 25\% & 30\% \\
Qwen 3.8 27B    & 3\% & 2\% & 1\% & 1\%  & 18\% & 59\% \\
\bottomrule
\end{tabular}
\end{table}

\section{Performance of Agents on Different Simulation Packages}
\begin{figure}[ht]
    \centering
    \includegraphics[width=1.0\linewidth]{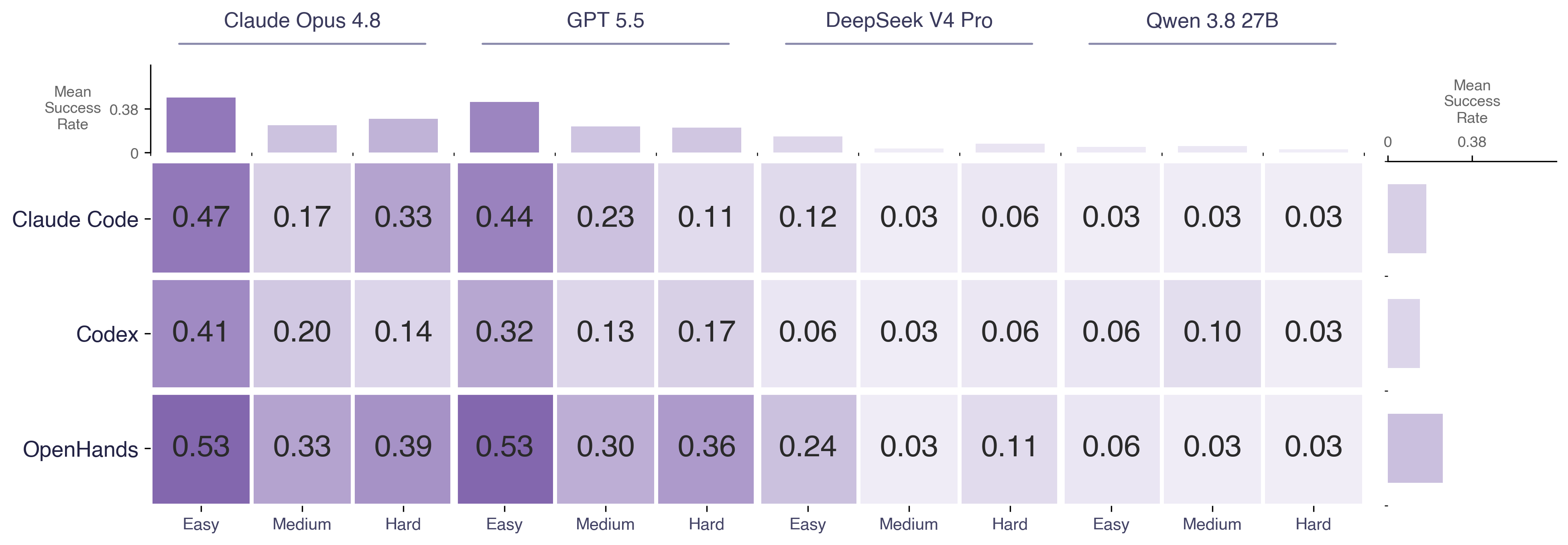}
    \caption{Full-success rate of each agent--model configuration by difficulty level on all \lammps{} tasks. Bars above and to the right give mean values over frameworks and models, respectively.}
    \label{fig:lammps_success_rate}
\end{figure}

\begin{figure}[ht]
    \centering
    \includegraphics[width=1.0\linewidth]{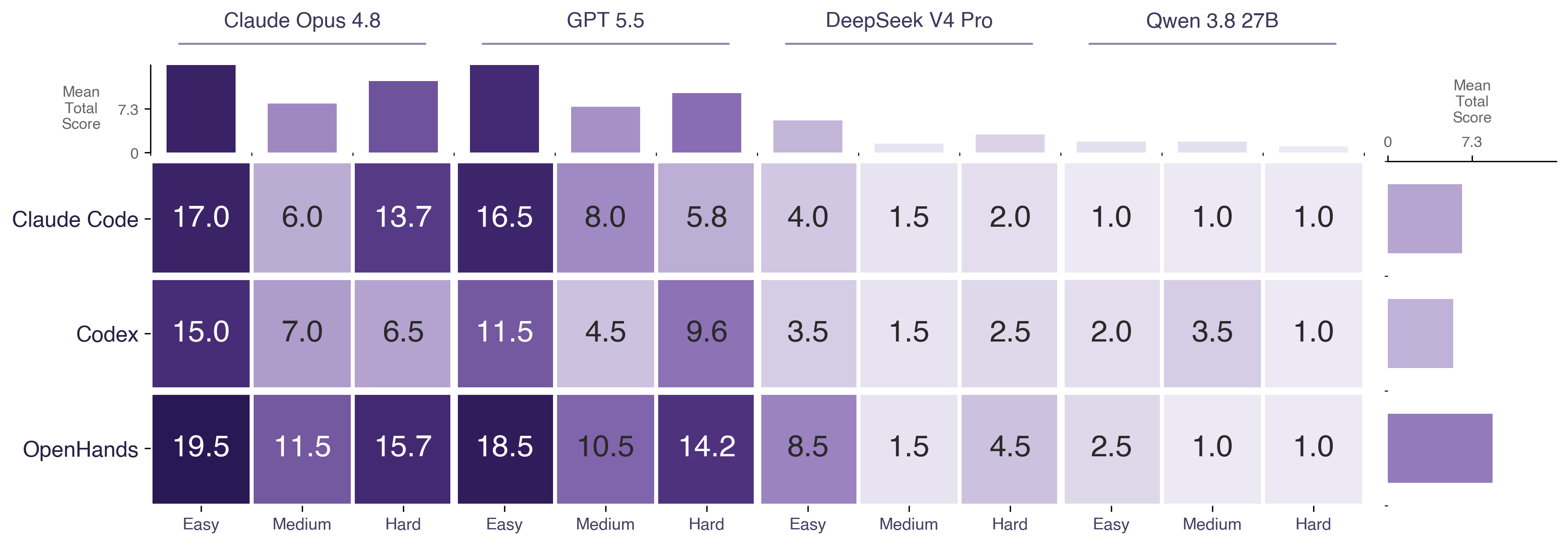}
    \caption{Total partial-credit score, the sum of per-problem scores $s$, of each agent--model configuration by difficulty level on all \lammps{} tasks. Bars above and to the right give means over frameworks and models, respectively.}
    \label{fig:lammps_total_score}
\end{figure}

\begin{figure}[ht]
    \centering
    \includegraphics[width=1.0\linewidth]{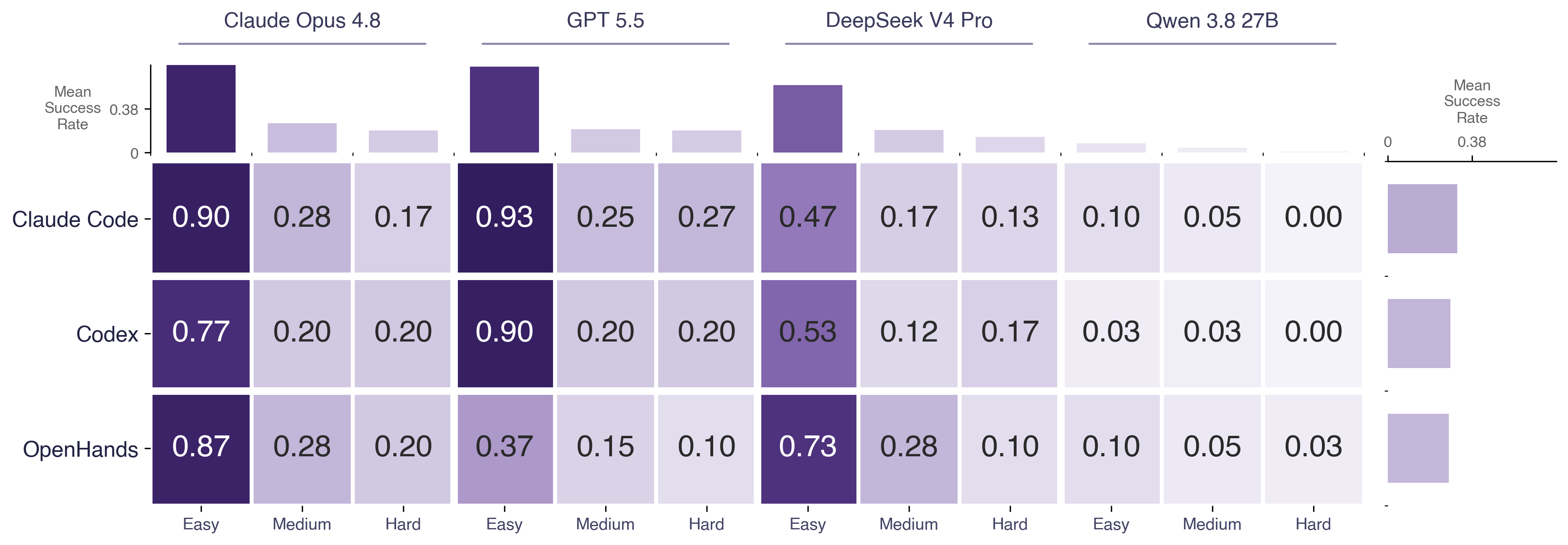}
    \caption{Full-success rate of each agent--model configuration by difficulty level on all \gromacs{} tasks. Bars above and to the right give mean values over frameworks and models, respectively.}
    \label{fig:gromacs_success_rate}
\end{figure}

\begin{figure}[ht]
    \centering
    \includegraphics[width=1.0\linewidth]{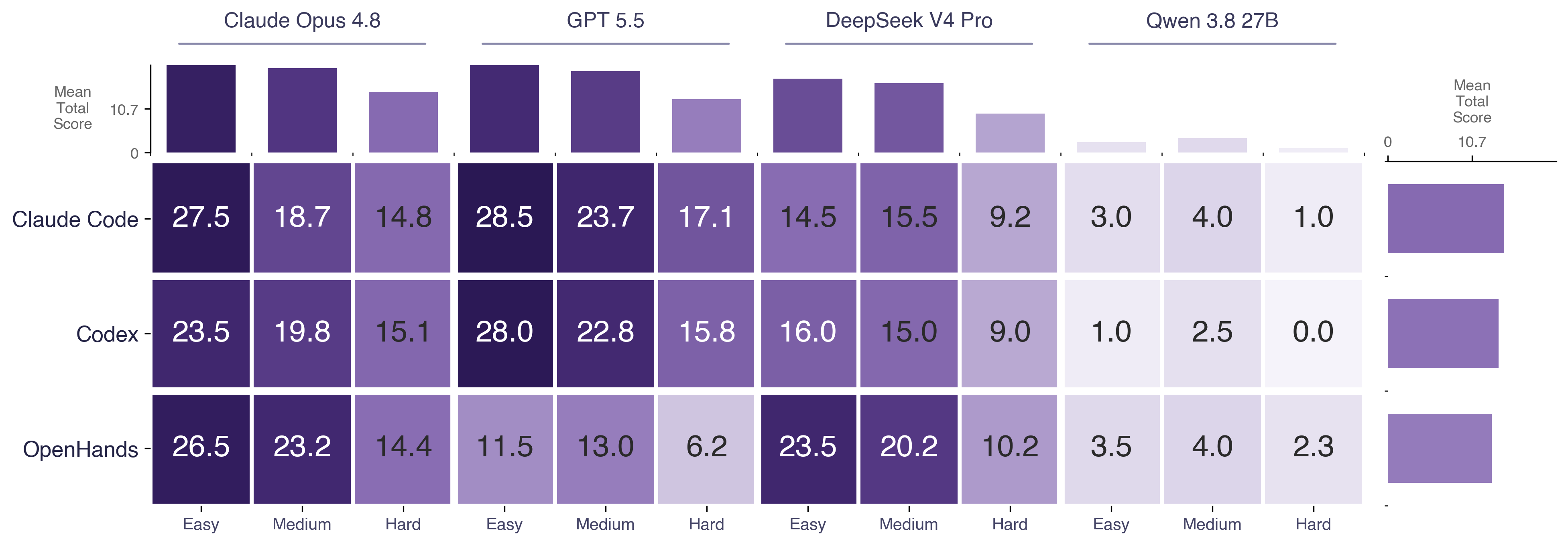}
    \caption{Total partial-credit score, the sum of per-problem scores $s$, of each agent--model configuration by difficulty level on all \gromacs{} tasks. Bars above and to the right give means over frameworks and models, respectively.}
    \label{fig:gromacs_total_score}
\end{figure}

\begin{figure}[ht]
    \centering
    \includegraphics[width=\linewidth]{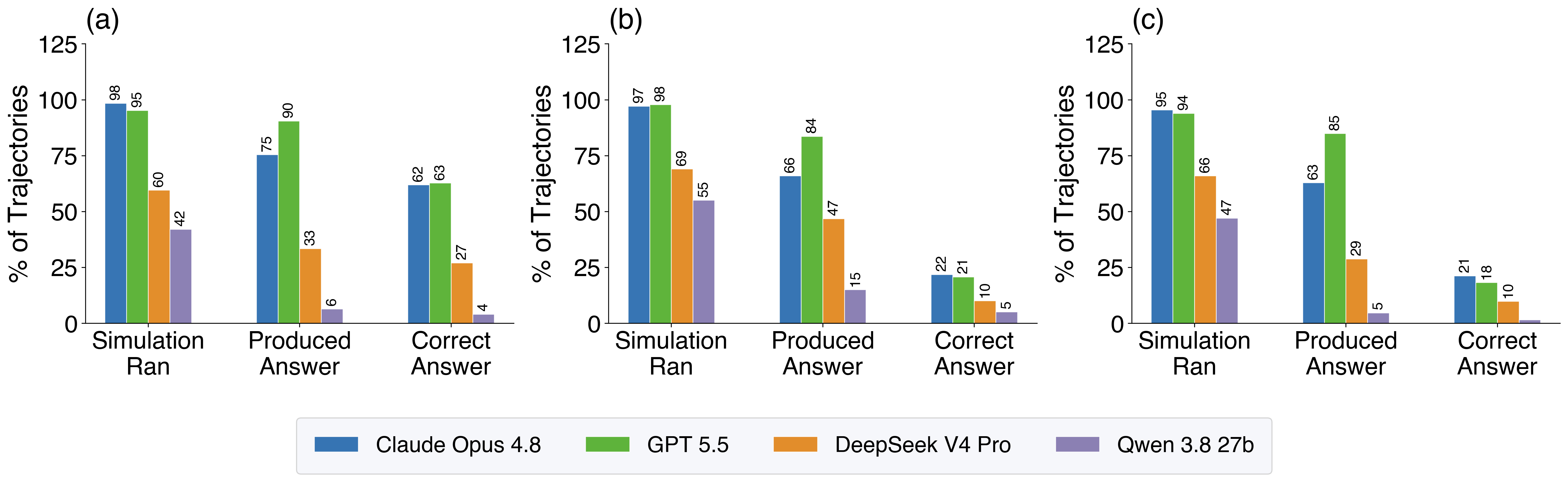}
    \caption{Task completion across difficulty levels, showing the percentage of trajectories reaching each of three stages: Simulation Ran (agent attempted execution within the time budget), Produced Answer (agent generated a \texttt{final\_answer.json} output), and Correct Answer (reported values within tolerance of reference). (a) Easy tasks. (b) Medium tasks. (c) Hard tasks. A consistent three-stage drop is observed across all difficulty levels and all agents.}
    \label{fig:task_completion_funnel}
\end{figure}

\section{Performance of all model--framework combinations}
\label{app:additional_results}

\begin{figure}[ht]
    \centering
    \includegraphics[width=1.0\linewidth]{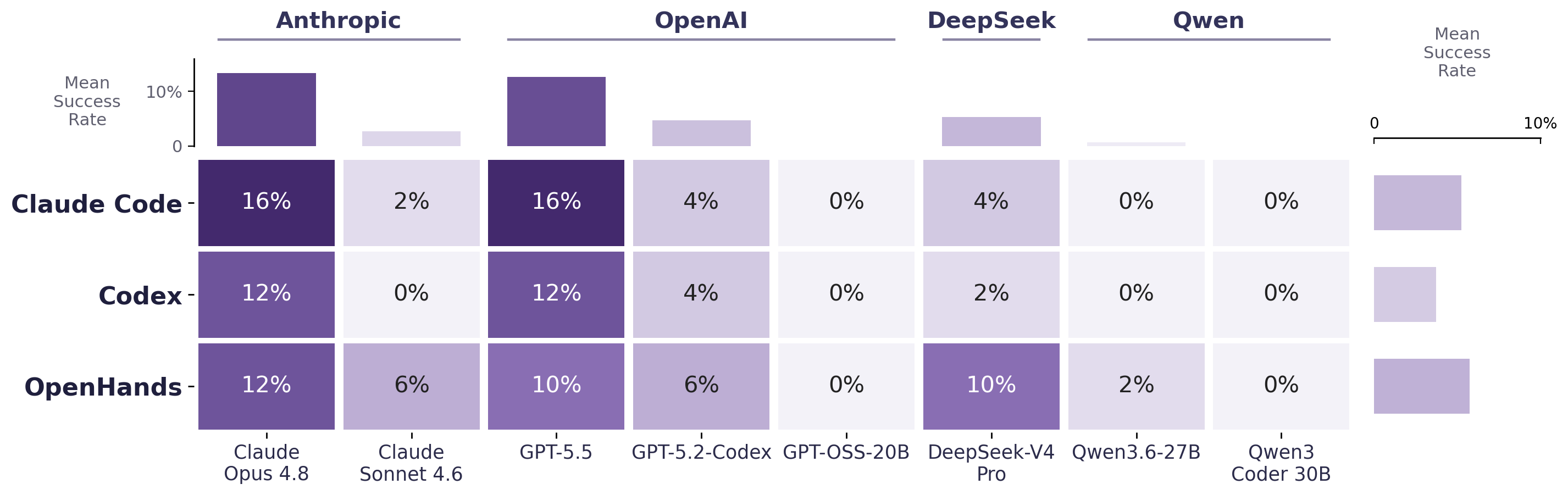}
    \caption{Full-success rate of every model--framework combination on a common subset of 50 Medium and Hard \mdgym{} problems. One problem corresponds to 2 percentage points.}
    \label{fig:all_combinations}
\end{figure}

To separate the contributions of the model and the agent framework, we evaluate every combination of eight LLMs and the three frameworks on a common subset of 50 problems sampled uniformly at random from the Medium and Hard levels of \mdgym{} (41 \lammps{} and 9 \gromacs{} problems). Besides the four models of the main evaluation, the subset includes Claude Sonnet 4.6, GPT 5.2 Codex, GPT OSS 20B, and Qwen3 Coder 30B. Figure~\ref{fig:all_combinations} reports the full-success rate of each combination. 

Consistent with the main results, we observe that the model determines performance far more strongly than the framework. Claude Opus 4.8 and GPT 5.5 solve 10--16\% of the problems under every framework, whereas GPT OSS 20B and Qwen3 Coder 30B solve none under any. Within each model family, the stronger model performs at least as well under every framework: Claude Opus 4.8 over Claude Sonnet 4.6, GPT 5.5 over GPT 5.2-Codex, and Qwen3.6-27B over Qwen3 Coder 30B. For a fixed proprietary model, the frameworks differ by at most 6 percentage points, or three problems, which lies within the sampling uncertainty of the subset. The native framework of a model is not consistently its best: Claude Sonnet 4.6 and GPT 5.2 Codex score highest under OpenHands, and GPT 5.5 under Claude Code, although none of these differences exceeds two problems.
These results agree with the evaluation on the full benchmark in Section~\ref{sec:results} and with the finding of \citet{rios2026ai}, across eight scientific domains, that the base model explains far more of the variance in agent performance than the scaffold.

\section{Statistical Significance of Results}
\label{appendix:stats}
We conduct four complementary statistical analyses on the per-problem partial credit scores ($s \in [0,1]$) and binary success rates across all 12 agent-model configurations (three harnesses × four models) reported in Section~\ref{sec:results}. All tests operate on per-problem scores and are computed using \texttt{scipy} and \texttt{statsmodels}. Raw scores are released alongside the benchmark to allow independent verification.
 
\rev{Because all 12 configurations attempt the same 200 problems, scores are not independent across configurations: a problem that is hard for one configuration tends to be hard for all. Tests that pool runs as independent observations ignore this structure and can misstate significance. For each claim we therefore report, alongside the pooled test, a \emph{paired} analysis that accounts for the shared problems. Configurations are compared on the same problem using the Friedman test and Wilcoxon signed-rank tests on scores, and Cochran's $Q$ on binary success, with Holm correction for multiple pairwise comparisons. Problem identity is modeled explicitly, either as a fixed factor in the ANOVA or as a random intercept in a linear mixed model. Confidence intervals on differences in success rate come from a problem-cluster bootstrap that resamples whole problems with replacement ($10^4$ resamples). Difficulty and engine are properties of the problem rather than of the configuration, so for these factors we average each problem's score over configurations and test on the 200 problem-level means, so that each problem counts once.}
 
\paragraph{Claim 1: Performance degrades with difficulty.}
We apply the Kruskal-Wallis test and Spearman rank correlation between difficulty level (coded ordinally as 1, 2, 3) and per-problem score, for each model (aggregated across harnesses) and combined overall. Degradation is highly significant for Claude Opus 4.8 ($H=47.19$, $p<0.001$; $\rho=-0.242$, $p<0.001$), GPT 5.5 ($H=32.12$, $p<0.001$; $\rho=-0.218$, $p<0.001$), and DeepSeek V4 Pro ($H=11.88$, $p=0.003$; $\rho=-0.140$, $p<0.001$). Qwen 3.8 27B shows a non-significant trend ($H=5.97$, $p=0.051$), reflecting a floor effect: scores are already near zero on easy tasks, leaving little statistical power to detect further decline. Combined across all 12 configurations, degradation is highly significant ($H=52.20$, $\rho=-0.144$, both $p<0.001$).
 
\rev{\textit{Paired analysis.} Testing on problem-level means ($n=200$) leaves the conclusion unchanged. The mean score falls from 0.425 at Easy to 0.286 at Medium and 0.243 at Hard ($H=19.51$, $p<0.001$; $\rho=-0.302$, $p<0.001$). The decline remains significant for Claude Opus 4.8 ($H=22.53$, $p<0.001$), GPT 5.5 ($H=19.54$, $p<0.001$), and DeepSeek V4 Pro ($H=6.12$, $p=0.047$). Most of the decline occurs between Easy and Medium; for Claude Opus 4.8 the Medium and Hard means are nearly equal (0.410 and 0.405). For Qwen 3.8 27B the levels differ ($H=7.18$, $p=0.028$), but the trend is not monotonic (means 0.068, 0.076, 0.032; $\rho=-0.109$, $p=0.126$).}
 
\paragraph{Claim 2: Model capability, not harness, is the dominant driver of performance.}
To directly address whether performance differences arise from the underlying model or from harness-specific engineering, we conduct a two-way ANOVA on per-problem scores with model, harness, and their interaction as factors, reporting the proportion of variance explained ($\eta^2$) by each. Model identity explains 16.4\% of total variance ($F=158.27$, $p<0.001$), harness identity explains only 0.14\% and is not statistically significant ($F=2.06$, $p=0.127$), and the model$\times$harness interaction explains a further 0.98\% ($F=4.74$, $p<0.001$). Model capability therefore accounts for over 100$\times$ more variance in performance than harness choice. We additionally verify rank consistency: the ranking of the four models by mean score is identical between Claude Code and Codex (Spearman $\rho=1.00$), and highly consistent between each of these and OpenHands ($\rho=0.80$ for both comparisons), confirming that stronger models remain the strongest performers regardless of harness.
 
\rev{\textit{Paired analysis.} Adding problem identity as a factor (Table~\ref{tab:anova_paired}) shows that it is the largest single source of variance (31.1\%). The model share is unchanged (16.4\%, $F=232.79$, $p<0.001$), and the harness main effect remains negligible in size (0.14\%), although it becomes marginally significant ($F=3.04$, $p=0.048$) once problem-level variance is removed from the residual. In a linear mixed model with a random intercept per problem and fixed effects for model and harness, neither harness coefficient differs from zero (Codex vs.\ Claude Code: $-0.023$, 95\% CI $[-0.054, +0.008]$, $p=0.149$; OpenHands vs.\ Claude Code: $+0.016$, $[-0.015, +0.047]$, $p=0.325$). The harness therefore has no consistent effect across models. Within a given model, however, the harness matters (Table~\ref{tab:harness_within}): paired Friedman tests across the three harnesses are significant for Claude Opus 4.8, GPT 5.5, and DeepSeek V4 Pro, but not for Qwen 3.8 27B. The direction of the effect depends on the model. OpenHands yields the highest score for Claude Opus 4.8 and DeepSeek V4 Pro but the lowest for GPT 5.5, which accounts for the significant interaction term.}
 
\paragraph{Claim 3: Stronger models significantly outperform weaker models.}
A Kruskal-Wallis test across the four models (pooling harnesses) confirms a highly significant overall difference ($H=425.54$, $p<0.001$). Mean scores are 0.492 (Claude Opus 4.8), 0.443 (GPT 5.5), 0.271 (DeepSeek V4 Pro), and 0.059 (Qwen 3.8 27B). Pairwise Mann-Whitney U tests show that all pairwise differences are significant at $p<0.001$ except Claude Opus 4.8 vs.\ GPT 5.5 ($p=0.052$); every other pair, including all comparisons against Qwen 3.8 27B, is significant at $p<0.001$.
 
\rev{\textit{Paired analysis.} Comparing models on the same problems (scores averaged over harnesses) confirms the overall difference (Friedman $\chi^2=280.5$, $p<0.001$), and all six pairwise Wilcoxon tests remain significant after Holm correction (Table~\ref{tab:model_paired}). Cochran's $Q$ on binary success is significant within each harness (Claude Code $Q=123.5$, Codex $Q=91.7$, OpenHands $Q=101.9$; all $p<0.001$). Pairing makes the Claude Opus 4.8 vs.\ GPT 5.5 difference significant ($+0.049$, Holm $p=0.013$; bootstrap success-rate difference $+4.8$ points, 95\% CI $[+0.8, +8.8]$). This advantage, however, comes entirely from OpenHands ($+0.184$, $p<0.001$); the two models do not differ under Claude Code ($-0.010$, $p=0.63$) or Codex ($-0.026$, $p=0.73$). We therefore interpret the results as three clearly separated tiers --- the two frontier models, DeepSeek V4 Pro, and Qwen 3.8 27B --- and do not claim a general ordering between Claude Opus 4.8 and GPT 5.5.}
 
\paragraph{Claim 4: The benchmark is genuinely hard.}
We report Wilson score 95\% confidence intervals on binary success rates for all 36 (harness, model, difficulty) combinations, summarized in Table~\ref{tab:wilson_ci_v2}. The single highest observed success rate --- Claude Opus 4.8 via OpenHands on easy tasks --- has an upper confidence bound of 78.8\%, and all configurations at hard difficulty have upper confidence bounds at or below 42.2\%, confirming that even the strongest agent-model pairs fail on the majority of hard problems and that this gap is not an artefact of sample size. Each interval is computed for a single configuration over distinct problems, so the pairing across configurations does not affect these intervals.

\begin{table}[ht]
\centering
\small
\caption{\rev{Variance decomposition of per-problem scores, without and with problem identity as a factor.}}
\label{tab:anova_paired}
\begin{tabular}{lcccccc}
\toprule
 & \multicolumn{3}{c}{\textbf{Without problem}} & \multicolumn{3}{c}{\textbf{With problem}} \\
\cmidrule(lr){2-4}\cmidrule(lr){5-7}
\textbf{Factor} & $F$ & $p$ & $\eta^2$ & $F$ & $p$ & $\eta^2$ \\
\midrule
Problem          & ---    & ---      & ---     & 6.65   & $<0.001$ & 31.1\% \\
Model            & 158.27 & $<0.001$ & 16.4\%  & 232.79 & $<0.001$ & 16.4\% \\
Harness          & 2.06   & 0.127    & 0.14\%  & 3.04   & 0.048    & 0.14\% \\
Model × Harness  & 4.74   & $<0.001$ & 0.98\%  & 6.97   & $<0.001$ & 0.98\% \\
Residual         & ---    & ---      & 82.5\%  & ---    & ---      & 51.4\% \\
\bottomrule
\end{tabular}
\end{table}
 
\begin{table}[ht]
\centering
\small
\caption{\rev{Effect of harness within each model, paired by problem. Mean partial-credit score and success rate per harness; Friedman test on scores and Cochran's $Q$ on binary success across the three harnesses.}}
\label{tab:harness_within}
\begin{tabular}{lccccc}
\toprule
\textbf{Model} & \textbf{Claude Code} & \textbf{Codex} & \textbf{OpenHands} & \textbf{Friedman $p$} & \textbf{Cochran's $Q$ $p$} \\
\midrule
Claude Opus 4.8 & 0.488 (38.0\%) & 0.435 (31.0\%) & 0.554 (42.5\%) & 0.003    & 0.002 \\
GPT 5.5         & 0.498 (36.0\%) & 0.461 (31.0\%) & 0.370 (30.0\%) & $<0.001$ & 0.153 \\
DeepSeek V4 Pro & 0.233 (16.0\%) & 0.237 (15.5\%) & 0.342 (24.5\%) & 0.001    & 0.003 \\
Qwen 3.8 27B    & 0.055 (4.0\%)  & 0.050 (4.0\%)  & 0.072 (5.0\%)  & 0.417    & 0.779 \\
\bottomrule
\end{tabular}
\end{table}

\begin{table}[ht]
\centering
\small
\caption{\rev{Pairwise model comparisons paired by problem (scores averaged over harnesses). $\Delta$ is the difference in mean partial-credit score; $p$ values are from Wilcoxon signed-rank tests with Holm correction. Success-rate differences are shown with 95\% problem-cluster bootstrap intervals.}}
\label{tab:model_paired}
\begin{tabular}{lccc}
\toprule
\textbf{Comparison} & $\Delta$ \textbf{score} & \textbf{Holm $p$} & $\Delta$ \textbf{success (95\% CI)} \\
\midrule
Claude Opus 4.8 vs.\ GPT 5.5         & $+0.049$ & 0.013    & $+4.8$  $[+0.8, +8.8]$ \\
Claude Opus 4.8 vs.\ DeepSeek V4 Pro & $+0.222$ & $<0.001$ & $+18.7$ $[+14.2, +23.3]$ \\
Claude Opus 4.8 vs.\ Qwen 3.8 27B    & $+0.434$ & $<0.001$ & $+33.0$ $[+27.5, +38.5]$ \\
GPT 5.5 vs.\ DeepSeek V4 Pro         & $+0.173$ & $<0.001$ & $+13.8$ $[+9.8, +18.0]$ \\
GPT 5.5 vs.\ Qwen 3.8 27B            & $+0.385$ & $<0.001$ & $+28.2$ $[+23.2, +33.2]$ \\
DeepSeek V4 Pro vs.\ Qwen 3.8 27B    & $+0.212$ & $<0.001$ & $+14.3$ $[+10.3, +18.3]$ \\
\bottomrule
\end{tabular}
\end{table}
 
\begin{table}[ht]
\centering
\small
\caption{\rev{Wilson 95\% confidence intervals on success rates, top and bottom 5 of 36 (harness, model, difficulty) configurations by success rate.}}
\label{tab:wilson_ci_v2}
\begin{tabular}{llcccc}
\toprule
\textbf{Harness} & \textbf{Model} & \textbf{Level} & \textbf{Successes/N} & \textbf{Rate} & \textbf{95\% CI} \\
\midrule
OpenHands   & Claude Opus 4.8 & Easy & 44/64 & 68.8\% & 56.6\%--78.8\% \\
Claude Code & Claude Opus 4.8 & Easy & 43/64 & 67.2\% & 55.0\%--77.4\% \\
Claude Code & GPT 5.5         & Easy & 43/64 & 67.2\% & 55.0\%--77.4\% \\
Codex       & GPT 5.5         & Easy & 38/64 & 59.4\% & 47.1\%--70.5\% \\
Codex       & Claude Opus 4.8 & Easy & 37/64 & 57.8\% & 45.6\%--69.1\% \\
\multicolumn{6}{c}{\textit{...}} \\
Claude Code & Qwen 3.8 27B    & Medium & 3/70 & 4.3\% & 1.5\%--11.9\% \\
OpenHands   & Qwen 3.8 27B    & Medium & 3/70 & 4.3\% & 1.5\%--11.9\% \\
OpenHands   & Qwen 3.8 27B    & Hard   & 2/66 & 3.0\% & 0.8\%--10.4\% \\
Claude Code & Qwen 3.8 27B    & Hard   & 1/66 & 1.5\% & 0.3\%--8.1\% \\
Codex       & Qwen 3.8 27B    & Hard   & 1/66 & 1.5\% & 0.3\%--8.1\% \\
\bottomrule
\end{tabular}
\end{table}

\subsection{Stability of Results}
\label{appendix:repeats}
\begin{table}[ht]
\centering
\small
\caption{\rev{Repeated runs on a 50-problem subset. Success rate is the mean $\pm$ sample standard deviation over three runs. ``Same verdict'' counts problems that are solved in all three runs or in none.}}
\label{tab:repeats}
\begin{tabular}{lcccc}
\toprule
\textbf{Configuration} & \textbf{Solved per run} & \textbf{Success rate} & \textbf{Same verdict} \\
\midrule
Claude Code + GPT 5.5         & 8 / 7 / 7 & 14.7\% $\pm$ 1.2\% & 40/50 \\
Claude Code + DeepSeek V4 Pro & 2 / 0 / 1 & 2.0\% $\pm$ 2.0\%  & 47/50 \\
\bottomrule
\end{tabular}
\end{table}

\rev{All results in the main text come from one run per configuration and problem. To test whether they are stable across runs, we ran two configurations, Claude Code with GPT 5.5 and Claude Code with DeepSeek V4 Pro, three times each on a subset of 50 problems. The two configurations share a framework and use models from different tiers, so the repeats test both the stability of each configuration's success rate and the stability of the gap between tiers. Table~\ref{tab:repeats} summarises the results.
The results are stable across runs. The number of problems solved changes by at most one or two between runs, a standard deviation of 1--2 percentage points. GPT 5.5 outperforms DeepSeek V4 Pro in every run, by 12--14 percentage points, so the separation between model tiers does not depend on a particular run. Most problems also receive the same verdict in every run: 40 of 50 for GPT 5.5 and 47 of 50 for DeepSeek V4 Pro. The remaining problems are solved in some runs but not others, but these changes largely cancel out, leaving the aggregate success rate nearly unchanged.}





\end{document}